\documentclass[12pt, draftclsnofoot, onecolumn]{IEEEtran}
\usepackage{amsmath,amssymb,amsfonts,mathrsfs,bm,amsthm}
\usepackage{amstext}
\usepackage{upgreek}
\usepackage{multicol}
\usepackage{indentfirst}
\usepackage{graphicx}
\usepackage{paralist}
\usepackage{multirow}
\usepackage{tabularx}
\usepackage[noadjust]{cite}

\usepackage{booktabs}
\usepackage{subfigure}
\usepackage{color}
\usepackage{textcomp}
\usepackage{url}
\usepackage{setspace}

\IEEEoverridecommandlockouts
\allowdisplaybreaks
\usepackage{algorithm,algorithmicx}
\floatname{algorithm}{\small \bf Algorithm}
\usepackage[noend]{algpseudocode}
\algrenewcommand{\algorithmiccomment}[1]{\hfill \# #1}
\newtheorem{corollary}{\bf Corollary}
\newtheorem{proposition}{\bf Proposition}

\newtheorem{remark}{\bf Remark}
\let\olddefinition\remark
\renewcommand{\remark}{\olddefinition\normalfont}

\usepackage{accents}

\allowdisplaybreaks
\definecolor{blue}{rgb}{0,0.24,0.54}
\begin{document}
\title{\huge Accelerating~Deep~Reinforcement~Learning With the Aid of Partial Model: Energy-Efficient~Predictive~Video~Streaming}
\author{
 	\IEEEauthorblockN{Dong Liu, Jianyu Zhao, Chenyang Yang, and Lajos Hanzo}\\
	\thanks{
	D. Liu and L. Hanzo are with the University of Southampton, Southampton SO17 1BJ, UK (email: \{d.liu, hanzo\}@soton.ac.uk).
	J. Zhao and C. Yang are with Beihang University, Beijing 100191, China (e-mail: jianyuzhao\_buaa@163.com, cyyang@buaa.edu.cn). This paper was presented in part at IEEE Globecom 2019 \cite{dongGC19}. The source code for reproducing the results of this paper is available at
	https://github.com/fluidy/twc2020.}
}
\maketitle

\vspace{-15mm}
\begin{abstract}
Predictive power allocation is conceived for energy-efficient video streaming over mobile networks using deep reinforcement learning. The goal is to minimize the accumulated energy consumption of each base station over a complete video streaming session under the constraint that avoids video playback interruptions. To handle the continuous state and action spaces, we resort to deep deterministic policy gradient (DDPG) algorithm for solving the formulated problem. In contrast to previous predictive power allocation policies that first predict future information with historical data and then optimize the power allocation based on the predicted information, the proposed policy operates in an on-line and end-to-end manner. By judiciously designing the action and state that only depend on slowly-varying average channel gains, we reduce the signaling overhead between the edge server and the base stations, and make it easier to learn a good policy. To further avoid playback interruption throughout the learning process and improve the convergence speed, we exploit the partially known model of the system dynamics by integrating the concepts of safety layer, post-decision state, and virtual experiences into the basic DDPG algorithm. Our simulation results show that the proposed policies converge to the optimal policy that is derived based on perfect large-scale channel prediction and outperform the first-predict-then-optimize policy in the presence of prediction errors. By harnessing the partially known model, the convergence speed can be dramatically improved.
\end{abstract}
\vspace{-5mm}
\begin{IEEEkeywords}
\vspace{-2mm}
Deep reinforcement learning, convergence speed, constraint, energy efficiency, video streaming
\end{IEEEkeywords}
\section{Introduction}
Mobile video traffic is expected to account for more than $79\%$ of the global mobile data by 2022, and video-on-demand (VoD) services represent the main contributor~\cite{index2017global}. Video streaming over cellular networks enables mobile users to watch the requested video while downloading. To avoid video stalling for a user experiencing hostile channel conditions, a base station (BS) can increase its transmit power for ensuring that the video segment is downloaded before being played. This, however, may cause a significant increase in energy consumption, hence degrading one of the most important design metrics of cellular networks, namely energy efficiency (EE).

The dynamic nature of wireless environment mainly owing to the user behavior, which has long been regarded as being random and remains unexploited in the design of wireless systems. However, with machine learning and the availability of big data, the user behavior becomes predictable to some degree and hence can be exploited for predictive resource allocation (PRA). For example, by predicting the user trajectory~\cite{zhang2018trajectory} and constructing radio coverage map \cite{kasparick2015kernel}, the future average channel gains in each \emph{time frame}\footnote{In this paper, a time frame refers to the duration of time, say one second, where the large-scale channel gain can be regarded as a constant, instead of the ``video frames" that compose a video segment.} (TF) can be predicted up to a minute-level time horizon. Based on the predicted future channel gains, the BS can proactively transmit more data in advance to the user's buffer during the instances of good channel conditions.

By harnessing various kinds of future information, PRA has been shown to provide a remarkable gain in improving the EE of mobile networks during video streaming ~\cite{tsilimantos2016anticipatory,abou2014energy,atawia2017robust,she2015context,mobility,GY18}. Assuming perfectly known future instantaneous channel gains, the trade-off between the required resources and the video stalling duration was investigated in \cite{tsilimantos2016anticipatory}. Assuming perfectly known future instantaneous data rates in each \emph{time slot}\footnote{A time slot typically has a duration of milliseconds, within which the small-scale channel fading can be regarded as a constant.} (TS), the total number of TS for video streaming was minimized in \cite{abou2014energy} to save energy. Considering that future data rates cannot be predicted without errors, the predicted data rate is modeled as random variables with known average values and bounded prediction errors in \cite{atawia2017robust} for optimizing PRA.  
Assuming known future average channel gains, the optimal PRA was derived in \cite{she2015context} for maximizing the EE of video streaming, and was extended to hybrid scenarios, where both real-time and VoD services coexist \cite{scy}.

To employ these optimized PRA policies, an immediate approach is to first predict the future information by machine learning, such as using a recurrent neural network (RNN)~\cite{zhang2018trajectory},  and then allocate radio resources by solving optimization problems based on the predicted information~\cite{mobility,GY18}. This approach is operated in four phases. In the first phase, a predictor (say for predicting the future average channel gains) is trained in an off-line manner using historical data \cite{zhang2018trajectory}. In the second phase, the data (say the locations along the user trajectory) for making a prediction is gathered after a user initiates a request. The third phase assigns radio resources to all the TFs or TSs in a prediction window at the start of the window. Finally, the BS allocates resources and transmits to the user in each TS according to the pre-assigned resources. However, such a \emph{first-predict-then-optimize} procedure is tedious and suffers from the following impediments:
\begin{itemize}
	\item When a user starts to play a video file, the predictor has to gather data for making the prediction. However, before any future resources are allocated based on the predicted information, the BS has to serve the user in a non-predictive manner. 
	\item The prediction accuracy degrades as the prediction horizon expands and the resultant PRA policy cannot be well-adapted to the dynamically fluctuating wireless environment. 
\end{itemize}

A natural question is: can we optimize PRA in an on-line and end-to-end manner?
Reinforcement learning (RL) \cite{sutton1998reinforcement} can be invoked for on-line learning. By further combining deep learning~\cite{lecun2015deep}, deep reinforcement learning (DRL) enables end-to-end learning of a policy that directly maps from observations to the desired action.
With the aid of the new paradigm of mobile edge computing (MEC)~\cite{hu2015mobile}, the training and inference of DRL algorithms can be implement at the wireless edge to address various challenging wireless tasks~\cite{DRL,zhao2019deep,zhang2019proactive,liu2019DRL}.

Against this background, we propose a DRL framework for optimizing predictive power allocation. We consider a scenario where users travel across multiple cells covered by a MEC server during video streaming. The objective is to minimize the accumulated energy consumption of each BS for the entire video streaming session under the quality-of-service (QoS) constraint that avoids video stalling. Such a problem can be formulated as a Markov decision process (MDP) with constraint. However, a straightforward implementation of standard DRL algorithm, such as deep deterministic policy gradient (DDPG)~\cite{DDPG}, would incur the following obstacles:

\begin{itemize}
	\item \emph{Highly dynamic small-scale channel fadings and excessive signaling overhead}: The straightforward way to implement DRL is to regard the instantaneous transmit power as the action. Then, the action and state should depend on the instantaneous channel gains, which however incurs millisecond-level information exchange between the MEC server and each BS, yielding excessive signaling overhead. Furthermore, this makes it hard for the agent to learn a good policy due to the highly dynamic small-scale channel fadings.
	\item \emph{Violation of constraint during learning}:  To guarantee the constraint, a heustric way is reward shaping by adding a penalty term into the reward when video stalling events occur, which however, introduces an extra hyper-parameter requiring sophistical tuning. Moreover, this approach cannot guarantee  constraint satisfaction during the entire learning process due to the random exploration action of DRL, which impairs the online performance.   
	\item \emph{Poor sample-efficiency}: General DRL algorithms are designed for model-free tasks, whose major obstacle in the real world implementation is their high sample complexity. Yet, for many wireless problems, a part of the dynamic model is known, which however, remains unexploited when implementing DRL.
\end{itemize}

In response to the issues with the \emph{first-predict-then-optimize} PRA and the obstacles faced by DRL, our major contributions can be summarized as follows:
\begin{itemize}
	\item {\it Simplify learning and avoid excessive signaling overhead by judiciously designing the state and action that only depend on slowly-varying information}: We first derive the optimal power allocation policy in closed-form for arbitrary average data rate in each TF by exploiting the knowledge concerning the distribution of small-scale fading. In this way, we can regard the average data rate as the action without loss of optimality, and hence the system's state only depends on the average channel gains. This avoids millisecond-level information exchange, and makes it easier for the agent to learn a good policy.
	\item {\it Satisfy the QoS constraint throughout the learning  process by adopting a safety layer:} Inspired by the idea of safe RL designed for the situations where the safety of the agent
	(say a robot) is particularly important \cite{dalal2018safe}, we design a safety layer for the actor network in DDPG for satisfying the QoS constraint during the entire learning process. This also avoids the tuning of an extra hyper-parameter compared with the reward shaping approach.
	In contrast to \cite{dalal2018safe} that is designed for completely unknown environments, the safety layer in this work is derived in closed-form by exploiting the partially known dynamic model.
	\item {\it Improve the sample-efficiency of DRL using post-decision state (PDS) and virtual experiences:} Inspired by the idea of introducing PDS to accelerate Q-learning by dividing the system's dynamics into known and unknown components \cite{mastronarde2011fast}, we integrate PDS into DDPG and propose an amalgamated PDS-DDPG algorithm, which significantly reduces the number of parameters to be learned by integrating available model into the neural networks (NNs). In contrast to \cite{mastronarde2011fast}, the proposed PDS-DDPG algorithm harnesses NNs and becomes eminently suitable for learning in continuous state and action spaces. Furthermore, we exploit the knowledge concerning the problem at hand that the unknown dynamics are independent of the known dynamics by generating virtual experiences based on historical data. By training with both virtual and real experiences, the sample efficiency can be boosted.
	\item {\it Outperform the first-predict-then-optimize approach for video streaming}: Our simulation results show that the proposed DRL-based policies converge to the optimal policy that is derived based on perfect channel prediction. Since DRL learns the transmission policy in an on-line manner, the proposed policies can adapt to the fluctuating wireless channel promptly and achieves lower energy consumption than the first-predict-then-optimized approach in the presence of prediction errors.
	\item {\it Provide the potential for wide application in wireless networks:} The ideas of integrating the PDS, safety layer and virtual experiences into DRL have the potential of applicability to many wireless tasks beyond video streaming. Moreover, those techniques can be implemented not only   upon DDPG, but also upon other advanced DRL algorithms, such as TD3~\cite{fujimoto2018addressing} and Q-Prop~\cite{gu2016q}.
\end{itemize}

\begin{table}[htbp]
	\centering
		\scriptsize
		\setlength{\extrarowheight}{1.5pt}
		\caption{Comparison with Existing Energy-Efficient Predictive Video Streaming Methods}
		\vspace{-2mm}
		\begin{tabular}{|m{3.3cm}|m{1.9cm}|m{1.8cm}|m{1.8cm}|m{1.8cm}|m{3.7cm}|}
			\hline
			\multirow{2}{*}{Method} & \multicolumn{4}{c|}{First-Predict-Then-Optimize Approach} & \multirow{2}{*}{Proposed DRL-Based Approach} \\
			\cline{2-5}
			& \multicolumn{1}{c|}{\cite{tsilimantos2016anticipatory}}   & \multicolumn{1}{c|}{\cite{abou2014energy}}   & \multicolumn{1}{c|}{\cite{atawia2017robust}}   & \multicolumn{1}{c|}{\cite{she2015context,scy}}  &  \\\hline
			 Future Information Required Within a Prediction Window& Future instantaneous channel gains  & Future instantaneous data rate  & Future average data rate  & Future average channel gains  & \multicolumn{1}{c|}{ Not required} \\\hline
			Optimization & \multicolumn{4}{c|}{Off-line optimization based on explicitly predicted information} & On-line and end-to-end learning \\\hline
		\end{tabular}%
		\label{tab:c1}%
\end{table}%

\vspace{-8mm}
\begin{table}[htbp]
	\centering
		\scriptsize
		\setlength{\extrarowheight}{1.5pt}
		\caption{Comparison with Existing RL/DRL Algorithms}
		\vspace{-2mm}
		\begin{tabular}{|l|r|r|r|m{2cm}|m{1.8cm}|}
			\hline
			RL/DRL algorithm & \parbox{1.5cm}{\centering DDPG~\cite{DDPG} \\ TD3~\cite{fujimoto2018addressing}} & Q-Prop~\cite{gu2016q} & PDS-Q-Learning~\cite{mastronarde2011fast} & Proposed PDS-DDPG/TD3 with Safety Layer &  Proposed PDS-Q-Prop with Safety Layer \\
			\hline
			Model Free & \multicolumn{1}{c|}{\checkmark}     & \multicolumn{1}{c|}{\checkmark}     &       &       &  \\\hline
			Exploit Available Model &       &       & \multicolumn{1}{c|}{\checkmark}    &\multicolumn{1}{c|}{\checkmark}      & \multicolumn{1}{c|}{\checkmark} \\\hline
			Satisfy Constraint During Learning &       &       &       & \multicolumn{1}{c|}{\checkmark}     & \multicolumn{1}{c|}{\checkmark} \\\hline
			Continuous Action & \multicolumn{1}{c|}{\checkmark}     & \multicolumn{1}{c|}{\checkmark}     &       & \multicolumn{1}{c|}{\checkmark}    & \multicolumn{1}{c|}{\checkmark} \\\hline
			Discrete Action &       &   \multicolumn{1}{c|}{\checkmark}    & \multicolumn{1}{c|}{\checkmark}     &       & \multicolumn{1}{c|}{\checkmark} \\\hline
		\end{tabular}%
		\label{tab:c2}%
\end{table}%
In Table~\ref{tab:c1} and \ref{tab:c2}, we summarize the comparison of our work with the most pertinent PRA and DRL algorithms, respectively.
The rest of the paper is organized as follows. In Section~II, we introduce the system model. In Section~III, we formulate the RL problem and solve it using  DDPG. In Section~IV, we exploit the partially known dynamics by integrating the concepts of safety layer, PDS and virtual experiences into the basic DDPG. Our simulations results are provided in Section~V, and finally, Section~VI concludes the paper.

\section{System Model}

In this section, we first introduce the notations to be used throughout the paper and the MEC-enabled network architecture to support DRL. Then, we describe the models related to wireless video streaming and formulate the energy consumption minimization problem.

\subsection{Baisc Notations in Reinforcement Learning and MEC-Enabled Network Architecture}

A standard RL problem can be formulated as an MDP, where an agent learns how to achieve a goal from its interactions with the environment in a sequence of discrete time steps $t = 1, 2,\cdots, T$ ~\cite{sutton1998reinforcement}. At each time step $t$, the agent observes the state $\mathbf s_t$ of the environment  and executes an action $ a_t$. Then, the agent receives a reward $r_t$ from the environment and transits into a new state $\mathbf s_{t+1}$. The interaction of the
agent with the environment is then captured by an experience vector $[\mathbf s_t,  a_t, r_t, \mathbf s_{t+1}]$.
The agent learns a policy from  its experiences for maximizing an expected return, which reflects the cumulative
reward received by the agent during the $T$-time-step episode. The policy determines which action should be executed in which state. The expected return is defined as $\mathbb E \left[\sum_{t=1}^{T-1} \gamma^{t-1} r_t\right]$, where $\gamma$ denotes the discount factor. All the notations to be used
throughout the paper are summarized in Table.~\ref{tab:notation}.

\begin{figure}[!htb]
	\vspace{-2mm}
	\centering
	\includegraphics[width=0.95\textwidth]{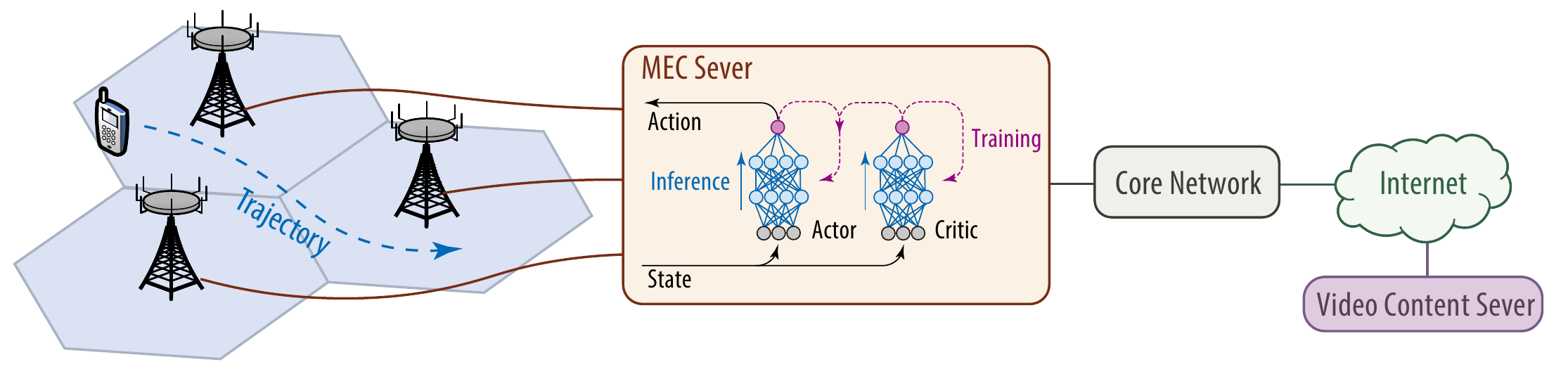}
	\vspace{-2mm}
	\caption{MEC-enabled cellular network.}
	\label{fig:arch}
	\vspace{-2mm}
\end{figure}
To implement DRL in cellular network, we consider a MEC-enabled network architecture as shown in Fig.~\ref{fig:arch}. The user may travel across multiple cells during the video streaming process and is associated the BS having the strongest average channel gain. 
The MEC server is located at an aggregation node (e.g., the serving gateway in the 4G LTE network~\cite{giust2018mec}) to support handover and acts as the DRL agent, which monitors and records the status of mobile users (say the channel conditions, buffer status) and BSs (say the consumed energy), learns the transmission policy and instructs the BSs to implement the learned policy.

\begin{table}
	\vspace{-5mm}
		\caption{Summary of Notations}
		\label{tab:notation}
		\scriptsize
		\setlength{\extrarowheight}{1.5pt}
		\vspace{-2mm}
		\begin{tabular}{|l|l||l|l|}\hline
	$\mathbf s_t$ & State of the $t$th TF	& ${\rm E}_1 (\cdot)$ & Exponential integral function \\\hline
	$a_t$ & Action of the $t$th TF &  $B_t$ & Amount of data in the user's buffer at the $t$th TF  \\\hline
	$r_t$ & Reward of received from transition $\mathbf s_t \to \mathbf s_{t+1}$ & 			$n_t$ & Index of the segment played in the $t$th TF \\\hline
	$\gamma$ & Discount factor & 			$l_t$ & Playback progress of the current video segment \\\hline
	$N_{\rm v}$& Number of segments in the video file  & 			$\eta_t$ & Download progress of the whole video file  \\\hline
	$S_n$ & Size of the $n$th video segment & 			$N_{\rm t}$ & Number of large-scale channel vectors in $\mathbf s_t$ \\\hline
	$L_{\rm v}$ & Number of TFs of each video segment  & 			$\lambda$ & Penalty coefficient \\\hline
	$\Delta T$ & Duration of each TF& 			$\mu(\cdot; \bm \theta_\mu)$ & Actor network with parameter $\bm \theta_\mu$ \\\hline
	$N_{\rm s}$ & Number of TS in each TF & 			$Q_\mu(\cdot)$ & Action-value function of policy $\mu$ \\\hline
	$\tau$ & Duration of each TS & 			$Q(\cdot; \bm \theta_Q)$ & Critic network with parameter $\bm \theta_Q$ \\\hline
	$\alpha_t$ & Large-scale channel gain from the nearest BS in the $t$th TF  & 			$\mathcal{B}$ & Indices set of a mini-batch \\\hline
	${\bm\alpha}_t$ & Large-scale channel gain vector in the $t$th TF & 			$\mathcal{D}$ & Replay buffer \\\hline
	$N_{\rm b}$ & Number of BSs in $\bm \alpha_t$ & 			$\delta$ & Learning rate \\\hline
	$g_{ti}$ & Small-scale channel gain in the $i$th TS of the $t$th TF & 			$\omega$ & Target network update rate \\\hline
	$R_{ti}$ & Instantaneous data rate in the $i$th TS of the $t$th TF  & 			$\mathcal{N}_t$ & Exploration noise \\\hline
	$\bar R_t$ & Average data rate of the $t$th TF  & 			$\bar p_t$ & Expectation of the transmit power in the $t$th TF\\\hline
	$p_{ti}$ & Transmit power in the $i$th TS of the $t$th TF  & 			$\bar p(\bar R_t, \alpha_t)$ & Function mapping from $\bar R_t$ and $\alpha_t$ to $\bar p_t$ \\\hline			
	$\sigma^2$ & Noise power  & 			$\tilde{\mathbf s}_t$ & Post-decision state \\\hline
	$W$ & Transmission bandwidth & 			$r_t^{\rm k}$ & Reward received from transition $\mathbf s_t \to \tilde{\mathbf s}$ \\\hline
	$\rho_{\rm E}$ & Power amplifier efficiency & 			$r_t^{\rm u}$ & Reward received from transition $\tilde{\mathbf s} \to \mathbf s_{t+1}$  \\\hline
	$P_{\rm c}$ & Circuit power consumption  & 			$\rho(\cdot | \cdot) $ & Conditional PDF \\\hline
	$P_{\max}$ & Maximum transmit power &		$V_{\mu}(\cdot)$  & PDS-value function of policy $\mu$ \\\hline
	$p^{\rm opt} (\cdot; \xi_t)$ & Optimal power allocation policy in the $t$th TF & 			$V(\cdot; \bm \theta_V)$ & PDS-value network with parameter $\bm \theta_V$  \\\hline
	$\xi_t$ & Optimal power allocation parameter & 				$\mathbf f_{\rm PDS}(\cdot)$ & PDS function  \\\hline		
	$\xi^{\rm opt} (\bar R_t)$ & Function mapping from $\bar R_t$ to $\xi_t$ & $\bm h$, $\mathcal{H}$ & Channel trace and channel trace buffer \\\hline
\end{tabular}
\end{table}

\subsection{Transmission and Channel Models}

\begin{figure}[!htb]
	\vspace{-2mm}
	\centering
	\includegraphics[width=0.85\textwidth]{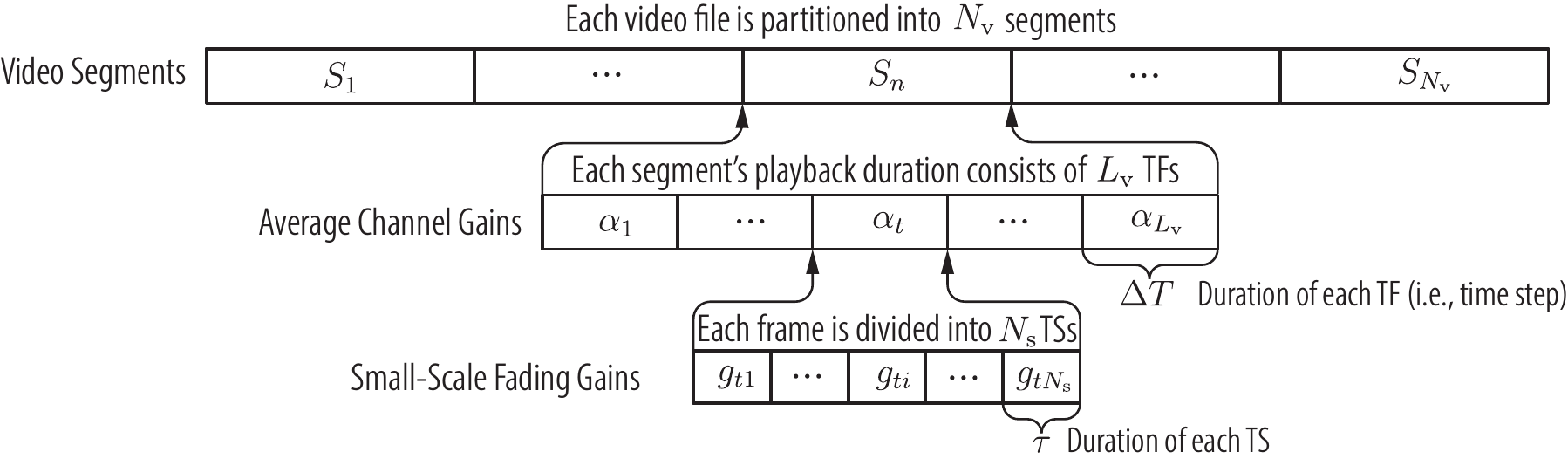}\vspace{-4mm}
	\caption{Video segment playback duration and channel variations.}
	\label{fig:time}
\end{figure}

Each video file is partitioned into $N_{\rm v}$ segments, each of which is the minimal unit for video playback. The playback duration of each segment is further partitioned into $L_{\rm v}$ TFs, each with a duration of $\Delta T$. Each TF is divided into $N_{\rm s}$ TSs, each having a duration of $\tau$, i.e., $\tau = \Delta T/N_{\rm s} $, as shown in Fig. \ref{fig:time}. The large-scale channel gains (i.e., average channel gains) are assumed to remain constant within each TF, and may change from one TF to another due to user mobility. The small-scale channel gains are assumed to remain constant within each TS, but they are independently and identically distributed (i.i.d.) among TSs.

Let $\alpha_t g_{ti}$ denote the instantaneous channel gain between a user and its associated BS in the $i$th TS of the $t$th TF, where $\alpha_t$ and $g_{ti}$ denote the large-scale channel gain and the small-scale fading gain, respectively. Upon assuming perfect capacity achieving coding, the instantaneous data rate in the $i$th TS of the $t$th TF can be expressed as
\begin{equation}
R_{ti} = W \log_2 \left(1 + \frac{\alpha_t g_{ti}}{\sigma^2} p_{ti}\right), \label{eqn:rate}
\end{equation}
where $W$ is the transmission bandwidth, $\sigma^2$ is the noise power, and $p_{ti}$ is the transmit power in the $i$th TS of the $t$th TF.

\subsection{Video Streaming and Power Consumption Model}

The video playback starts after a user has received the first video segment. To avoid stalling, each segment should be downloaded to the user's buffer before being played. We assume that the buffer capacity is higher than the video file size, which is reasonable for contemporary mobile devices. Hence, no buffer overflow is considered. Then, the following QoS constraint should be satisfied
\begin{equation}
\sum_{n=1}^{m} \sum_{t=(n-1)L_{\rm v} + 1}^{nL_{\rm v}}  \sum_{i=1}^{N_{\rm s}} \tau R_{ti}\geq \sum_{n=2}^{m+1}  S_n, ~ m = 1,\cdots,N_{\rm v} -1, \label{eqn:QoS}
\end{equation}
where $\sum_{t=(n-1)L_{\rm v} + 1}^{nL_{\rm v}} \sum_{i=1}^{N_{\rm s}} \tau R_{ti}$ is the amount of data transmitted to the user during the playback of the $n$th segment, $S_n$ is the size of the segment that is known after the user issues a request.

The energy consumed by a BS for video transmission during the $t$th TF is modeled as \cite{energy}
\begin{equation}\label{energy-con}
E_t = \frac{1}{\rho_{\rm E}} \sum_{i=1}^{N_{\rm s}} \tau p_{ti} + \Delta T P_{\rm c},
\end{equation}
where $\rho_{\rm E}$ reflects the power-efficiency of the power amplifier, and $P_{\rm c}$ is the power consumed by the baseband and radio frequency circuits as well as by the cooling and power
supply.

To find the best power allocation among TSs that minimizes the average energy consumption subject to the QoS and maximal power constraints, the problem can be formulated as
\begin{subequations} \label{eqn:p0}
	\begin{align}
{\sf P1}: \quad	\min_{\{p_{ti}\}}~& \mathbb{E}_{\alpha_t, g} \left[ \sum_{t=1}^{(N_{\rm v} - 1)L_{\rm v}} \left(\frac{1}{\rho_{\rm E}} \sum_{i=1}^{N_{\rm s}} \tau p_{ti} + \Delta T P_{\rm c} \right)\right] \label{eqn:obj}\\
	s.t.~ & \sum_{n=1}^{m} \sum_{t=(n-1)L_{\rm v} + 1}^{nL_{\rm v}}  \sum_{i=1}^{N_{\rm s}} \tau R_{ti}\geq \sum_{n=2}^{m+1}  S_n, ~ m = 1,\cdots,N_{\rm v} -1 \\
	& p_{ti} \leq P_{\max} ,~ \forall ~t, i,
	\end{align}
\end{subequations}
where $\mathbb{E}_{\alpha_t, g}[\cdot]$ denotes the expectation taken over both the large-scale and small-scale fadings, while $P_{\max}$ is the maximal transmit power of each BS. The distribution of $\alpha_t$ depends on the user's mobility pattern and $R_{ti}$ in the QoS constraint depends on $g_{ti}$ and $\alpha_t$, all of which are unknown in advance. Without future channel information, it is impossible to solve problem $\sf P1$ at the beginning of video streaming. In the sequel, we resort to RL to find the solution in an end-to-end manner.

\section{Energy-Saving Power Allocation Based on DDPG}
In this section, we establish a RL framework for $\sf P1$  and propose a policy learning algorithm.

\subsection{Reinforcement Learning Framework}
To implement RL, it is important to design the state, action and reward properly. As we mentioned in the introduction, the straightforward way of implementing RL by regarding the instantaneous power $p_{ti}$ as the action will yield excessive signaling overhead and will make it hard for the agent to learn a good policy due to the highly dynamic small-scale fading.
In fact, the small-scale fading gain $g_{ti}$ can be regarded as a multiplicative impairment imposed on $\alpha_t$ and hence $\alpha_tg_{ti}$ has a much higher dynamic range than $\alpha_t$. This inspires us to find the action and the state that only depend on $\alpha_t$ in the following.
\subsubsection{Action}

In practice, it is not hard to measure the distribution of small-scale fading. Based on the distribution, we can first derive the optimal power allocation policy for each TF to minimize the average energy consumed in the TF to achieve an arbitrarily given average data rate. Then, by optimizing the average rate for each TF, we can obtain the optimal power allocation for the whole video streaming session to minimize the overall energy consumption. This suggests that we can select the average data rate of each TF as the action. In this way, the action and state for the RL agent are independent of $g_{ti}$.

Based on \eqref{energy-con}, the average energy consumption in the $t$th TF can be expressed as
\begin{equation}\label{aver-enegry}
\bar E_t = \mathbb{E}_{g}\left[\frac{1}{\rho_{\rm E}} \sum_{i=1}^{N_{\rm s}} \tau p_{ti} \right] + \Delta T P_{\rm c},
\end{equation}
where $\mathbb{E}_{g}[\cdot]$ denotes the expectation taken over small-scale fading.
Then, the objective function of problem $\sf P1$ can be rewritten as  $\mathbb{E}_{\alpha_t} \left[ \sum_{t=1}^{(N_{\rm v}- 1)L_{\rm v} } \bar E_t \right]$. For the $t$th TF, to achieve an arbitrarily given average rate $\bar R_t$, the optimal power allocation minimizing the average energy consumption in the $t$th TF can be found by solving,
\begin{subequations}
\begin{align}
{\sf P2}:\forall t, \quad  \min_{\{p_{ti}\}}~&  \bar E_t \\
s.t. ~& \mathbb{E}_{g}\left[W\log_2\left(1 + \frac{\alpha_t}{\sigma^2}p_{ti} g_{ti}\right)\right] = \bar R_t \\
& 0 \leq p_{ti} \leq P_{\max}, ~\forall i.
\end{align}
\end{subequations}
The optimal solution of $\sf P2$ is given by the following proposition.
\begin{proposition}
The optimal power allocation policy in the $t$th TF is
\begin{align}
p^{\rm opt}(\alpha_t g_{ti}; \xi_t) = \left\{\begin{array}{ll}
0, ~ \alpha_tg_{ti} \leq \frac{\sigma^2}{ \xi_t} \\
\xi_t - \frac{\sigma^2}{\alpha_tg_{ti}},~ \frac{\sigma^2}{ \xi_t} < \alpha_tg_{ti} < \frac{\sigma^2}{\xi_t - P_{\max}} \\
P_{\max}, ~\alpha_tg_{ti}\geq  \frac{\sigma^2}{\xi_t - P_{\max}}, 
\end{array}
\right. \label{eqn:popt}
\end{align}
where the parameter $\xi_t$ can be obtained by solving the following equation
\begin{equation}
\bar R_t = \int_{0}^{\infty} W \log_2\left( 1 + \frac{\alpha_t}{\sigma^2} p^{\rm opt}(\alpha_t g, \xi_t) g \right) \rho(g){\rm d}g \label{eqn:relation}
\end{equation}
via bisection search, and $\rho(g)$ denotes the probability density function (PDF) of $g_{ti}$.
\end{proposition}
\begin{IEEEproof}
Problem $\sf P2$ can be rewritten as a functional optimization problem. Then, from the Karush-Kuhn-Tucker (KKT) conditions of the functional optimization problem, the optimal power allocation policy can be obtained. Detailed proof is provided in Appendix A.
\end{IEEEproof}

Let the function $\xi^{\rm opt}(\bar R_t)$ denote the relationship between $\xi_t$ and $\bar R_t$ found from \eqref{eqn:relation}, i.e., $\xi_t \triangleq \xi^{\rm opt}(\bar R_t)$, whose expression can be obtained for a special case in the following corollary.
\begin{corollary}
For Rayleigh fading and a large value of $P_{\max}$, we have
\begin{equation}
\xi^{\rm opt}(\bar R_t) = \frac{\sigma^2}{\alpha_t}\left[{\rm E}_1^{-1} \left(\frac{\bar R_t \ln 2}{W} \right)\right]^{-1}, \label{eqn:xiopt}
\end{equation}
where  ${\rm E}_1^{-1}(x)$ denotes the inverse function of the exponential integral function ${\rm E}_1(x) \triangleq \int_{x}^{\infty}\frac{e^-t}{t} dt$.
\end{corollary}
\begin{IEEEproof}
In this case, we have $\rho(g) = e^{-g}$ and the maximal transmit power constraint can be safely neglected. Corollary 1 can be then obtained from~\eqref{eqn:relation}.
\end{IEEEproof}

\begin{remark}
Proposition 1 and Corollary 1 provide the optimal optimal power allocation policy in the $t$th TF for an arbitrarily given average data rate $\bar R_t$. This means that the original problem $\sf P1$ can be solved equivalently by first optimizing the average rate for each TF, i.e., $\{\bar R_t\}_{t=1,\cdots,(N_{\rm v} - 1)L_{\rm v}}$, and then obtaining the optimal power allocation for each TF using \eqref{eqn:popt}.
\end{remark}

As suggested by Remark 1, the MEC server only has to decide the {\bf action} as
\begin{equation}
a_t = \bar R_t.
\end{equation}
Upon determining the action, the MEC server can compute $\xi_t$ by bisection search based on \eqref{eqn:relation} in general cases or by \eqref{eqn:xiopt} for the special case given in Corollary~1, followed by sending $\xi_t$ to the specific BS that the user is associated with. According to $p_{ti} = p^{\rm opt} (\alpha_tg_{ti}; \xi_t)$, the BS can adjust the transmit power in each TS of the $t$th TF.
In this way, the agent interacts with the environment on a frame-by-frame basis (i.e., the time step is set as a TF on a second-level timescale as shown in Fig.~\ref{fig:time}) and the communication overhead of the MEC server can be reduced, while the BS can adjust the transmit power for each TS on a millisecond-level timescale.
\subsubsection{State}
Since the average power consumed by video transmission depends on the large-scale channel gain, $\alpha_t$ should be included into the state. To help the agent implicitly learns the user's mobility pattern, the state should also include the channel gains in the past $N_t$ TFs. Since different locations of a user may result in the same large-scale channel gain between the user and its associated BS, we further include the large-scale channel gains between the user and $(N_b-1)$ adjacent BSs. Let us now define a channel vector $\bm \alpha_{t} \triangleq [\alpha_{1, t}, \cdots, \alpha_{N_b, t}]$, where $\alpha_{n,t}$ is the large-scale channel gain between the user and the BS with the $n$th largest large-scale channel gain, and $\alpha_{1,t}$  is the large-scale channel gain between the user device and its associated BS (i.e., $\alpha_{1,t} = \alpha_t$). To meet the QoS requirement, the buffer status at the user should also be incorporated into the state. Let $B_t$ denote the amount of data remaining in the user's buffer at the $t$th TF. The transition of $B_{t}$ obeys:
\begin{equation}
B_{t+1} = B_t + \sum_{i=1}^{N_{\rm s}}\tau R_{ti} - I (l_t = L_{\rm v})S_{n_t},  \label{eqn:B}
\end{equation}
where $\sum_{i=1}^{N_{\rm s}}\tau R_{ti}$ is the amount of data transmitted during the $t$th TF,  $l_t \in [0, L_{\rm v}]$  denotes the number of TFs that the current segment has been played without stalling (which reflects the playback progress of the current segment) by the end of the $t$th TF, $I(\cdot)$ is an indicator function that equals $1$ if its argument is true and $0$ otherwise. When $l_t = L_{\rm v}$, the segment has been completely played by the end of the $t$th TF and the next segment is expected to be played in the $(t+1)$th TF. We use $n_t$ to denote the index of the segment played in the $t$th TF and the size of the $n_t$th segment is denoted by $S_{n_t}$. The last term $-I (l_t = L_{\rm v})S_{n_t} $ of \eqref{eqn:B} means that the $n_t$th segment will be removed from the buffer when its playback is completed. The evolution of the playback process obeys:
\begin{equation}
l_{t+1} =\left\{\begin{array}{ll}
{\rm mod}(l_t, L_{\rm v}), &~\text{if}~S_{n_{t+1}} > B_{t+1}\text {, i.e., video stalls}  \\
{\rm mod}(l_t, L_{\rm v}) + 1, &~\text{otherwise.}  \\
\end{array}
\right.  \label{eqn:lt}
\end{equation}

As shown in \eqref{eqn:B}, both $l_t$ and $S_{n_t}$ affect the transition of $B_t$ to $B_{t+1}$ and hence they should be included into the state. Moreover, since the portion of video file having been downloaded affects the termination of the episode, we include the ratio of the accumulated downloaded bits to the whole video file size
\begin{equation}
\eta_t = \frac{\sum_{j=1}^{t} \sum_{i=1}^{N_{\rm s}}\tau R_{ji}}{\sum_{n=1}^{N_{\rm v}}S_n} \label{eqn:download}
\end{equation}
into the state vector to reflect the downloading progress of the video file. Finally, the {\bf state} vector is designed as
\begin{equation}
\mathbf s_t = [B_t, S_{n_t}, l_t, \eta_t, \bm \alpha_{t}, \bm \alpha_{t-1}\cdots, \bm \alpha_{t-N_t}]. \label{eqn:st}
\end{equation}

\subsubsection{Reward}
The {\bf reward} for the agent is designed as
\begin{equation}
r_t = - \sum_{i=1}^{N_{\rm s}} \tau p_{ti} - \lambda \max\{S_{n_{t+1}} - B_{t+1} , 0\}, \label{eqn:reward}
\end{equation}
where $\sum_{i=1}^{N_{\rm s}} \tau p_{ti}$ is the transmit energy consumed in the $t$th TF, while $n_{t+1}$ is the index of the segment to be played in the next TF. The term $-\lambda \max\{S_{n_{t+1}} - B_{t+1} , 0\} $ imposes a penalty on the reward, when the amount of data in the user's buffer is less than the size of the segment to be played (i.e., when playback stalls). $\lambda$ is the penalty coefficient.

\subsection{Transmission Policy Based on DDPG}
The state vector defined in \eqref{eqn:st} lies in the continuous space. If $\mathbf s_t $ is discretized, then the number of possible states will be huge due to the combinatorial elements of $\mathbf s_t$, and the tabular-based RL (such as Q-learning \cite{sutton1998reinforcement}) encounters the curse of dimensionality. Moreover, the action $\bar R_t$ also lies in the continuous space. Value-based DRL methods such as deep Q-networks (DQNs)~\cite{mnih2015human} are designed for discrete action space and hence they are not suitable. DDPG \cite{DDPG} is designed based on the actor-critic architecture and it is able to learn a continuous policy. In contrast to other actor-critic-based RL algorithms that employ a stochastic policy gradient, DDPG employs a deterministic policy gradient so that the gradient of the policy can be estimated more efficiently~\cite{DDPG}. Therefore, we apply DDPG for solving the RL problem.

\begin{figure}[!htb]
	\vspace{-2mm}
	\centering	
	\subfigure[Actor network $\mu(\mathbf s_t; \bm \theta_{\mu}\!)$ and power allocation policy $p^{\rm opt}\!(\alpha_{t}g_{ti}; \xi_t\!)$]{
		\label{fig:ddpg-a} 
		\includegraphics[height=0.19\textwidth]{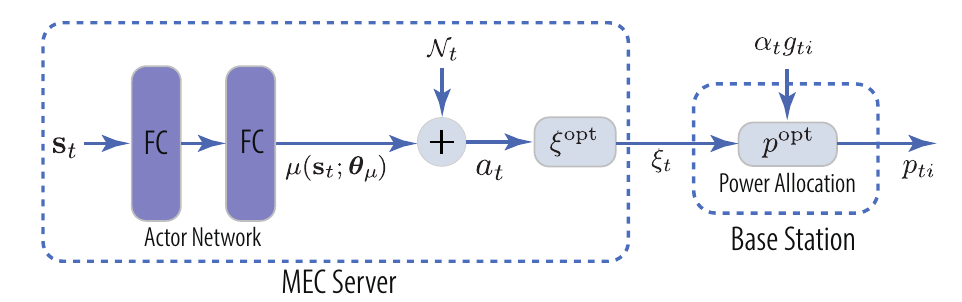}}
	\subfigure[Critic network $Q(\mathbf s_t, a_t; \bm \theta_Q)$]{
		\label{fig:ddpg-c} 
		\includegraphics[height=0.19\textwidth]{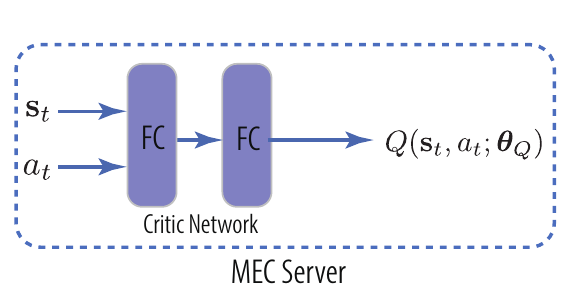}}
	\caption{Architecture of the actor and critic networks for DDPG. ``FC" denotes the fully-connected layer.}
	\label{fig:ddpg}
		\vspace{-2mm}
\end{figure}

DDPG maintains two NNs, namely the actor network $\mu (\mathbf s_t;\bm \theta_{\mu})$ and the critic network $Q(\mathbf s_t, a_t; \bm \theta_Q)$. The NNs' architecture in our framework is shown in Fig. \ref{fig:ddpg}. The actor network specifies the current policy by deterministically mapping each state into a specific continuous action. The output of the actor network is then used for computing the parameter $\xi_t = \xi^{\rm opt} (\bar R_t)$ according to Proposition 1 or Corollary 1. Upon receiving $\xi_t$ from the MEC server, the BS controls the transmit power at each TS within the $t$th TF according to the policy $p^{\rm opt} (\alpha_tg_{ti}; \xi_t)$ based on the current instantaneous channel gain $\alpha_t g_{ti}$. The critic network is used for approximating the \emph{action-value function}, $Q_{\mu}(\mathbf s_t, a_t) \triangleq \mathbb E \left[ \sum_{i=t}^{T} \gamma^{i-t}  r_i \big| \mathbf s_t , a_t, \mu\right]$, which is the expected return achieved by policy $\mu$, when taking action $a_t$ under state~$\mathbf s_t$.

During the interactions with the environment, the MEC server collects the experience $\mathbf e_t =  [\mathbf s_t,  a_t, r_t, \mathbf s_{t+1}]$ from the BSs in a database $\mathcal{D}$. A mini-batch of the experiences $\mathcal{B}$ is sampled from $\mathcal{D}$ to update the network parameters in each TF. The parameters of the critic network are updated using the batch gradient descent as
\begin{equation}
\bm \theta_{Q} \leftarrow \bm \theta_{Q} - \frac{\delta_Q}{|\mathcal{B}|} \nabla_{\bm \theta_{Q}} \sum_{j\in \mathcal{B}} \left[ y_j - Q(\mathbf s_j, a_j;\bm  \theta_Q) \right]^2, \label{eqn:Q}
\end{equation}
where $\delta_Q$ is the learning rate of the critic network, while we have $y_j = r_j$ if all the segments have been transmitted to the user and $y_j = r_j + \gamma Q'(\mathbf s_{j+1}, \mu'(\mathbf s_{j+1}; \bm \theta_{\mu}'); \bm \theta_{Q}')$ otherwise. Furthermore, $Q'(\mathbf s, a; \bm\theta_{Q}')$ and $\mu'(\mathbf s;\bm  \theta_{\mu}')$ are the target critic network and target actor network, respectively, which have the same structure as $Q(\cdot)$ and $\mu(\cdot)$. They are respectively updated by $\bm \theta_Q' \leftarrow  \omega \bm\theta_Q +  (1 - \omega) \bm \theta_{Q}'$ and $\bm \theta_\mu' \leftarrow  \omega \bm \theta_\mu +  (1 - \omega) \bm \theta_\mu'$ using a very small value of $\omega$ to stabilize the learning~\cite{DDPG}.

The parameters of the actor network are updated using the sampled policy gradient as
\begin{equation}
\bm \theta_{\mu} \leftarrow \bm \theta_{\mu} +  \frac{\delta_\mu}{|\mathcal{B}|} \sum_{j\in\mathcal{B}} \nabla_{a} Q(\mathbf s_j, a;\bm \theta_Q)|_{a = \mu(s_j;\bm \theta_\mu)} \nabla_{\bm \theta_\mu} \mu (\mathbf s_j;\bm \theta_{\mu}), \label{eqn:mu}
\end{equation}
where $\delta_\mu$ is the learning rate of the actor network.

To find the optimal policy, the agent has to explore the action space during the interactions with the environment. A noise term is added to the output of the actor network \cite{DDPG} to encourage exploration, which is formulated as $a_t = \mu(\mathbf s_t; \bm \theta_{\mu}) + \mathcal{N}_t$. The detailed procedure of learning the transmission policy is summarized in Algorithm 1.

\begin{algorithm}[!htb]
	\linespread{1}
	\caption{\small Learning Transmission Policy Based on DDPG}\
	\label{alg1}
	\small
	\begin{algorithmic}[1]
		\State Initialize critic network $Q(\mathbf s, a;\bm \theta_Q)$ and actor network $\mu (\mathbf s;\bm \theta_{\mu})$ with random weights $\bm \theta_{Q}$, $\bm \theta_{\mu}$.
		\State Initialize target networks $Q'$ and $\mu'$ with weights $\bm \theta_{Q}'	\leftarrow \bm \theta_{Q}$, $\bm \theta_{\mu}' \leftarrow \bm \theta_{\mu}$.
		\For{episode $= 1, 2, \cdots$}
		\For{TF $t = 1, 2, \cdots $}
		\State 	Observe state $\mathbf s_t$, select action $a_t = \mu(\mathbf s_t; \bm \theta_{\mu}) + \mathcal{N}_t$, set $\bar {R}_t = a_t$ and $\xi_t = \xi^{\rm opt}(\bar R_t)$.
		\For{TS $i = 1, \cdots, N_{\rm s}$}
		\State   Allocate transmit power according to \eqref{eqn:popt}.
		\EndFor
		\State Observe reward $r_t$ and new state $\mathbf s_{t+1}$.
		\State Store experience $[\mathbf{s}_t, a_t, r_t, \mathbf s_{t+1} ]$  in $\mathcal D$ and randomly sample a mini-batch of experiences $\mathcal{B}$ from~$\mathcal{D}$.
		\State Update the actor and critic networks according to \eqref{eqn:Q} and~\eqref{eqn:mu}, respectively.
		\State Update the target networks: $\bm \theta_{\mu}' \leftarrow \omega {\bm\theta}_{\mu} + (1-\omega) {\bm \theta}_{\mu}'$, ${\bm \theta}_{Q}' \leftarrow \omega {\bm\theta}_{Q} + (1-\omega) {\bm \theta}_{Q}'$.
		\If{all video segments have been transmitted to the user}
		\State {\bf break}
		\EndIf
		\EndFor
		\EndFor
	\end{algorithmic}
\end{algorithm}

\section{PDS-DDPG with Safety Layer and Virtual Experience}
In this section, we exploit the partial model concerning the dynamics of the system for satisfying the QoS constraint and improving the learning efficiency of the DDPG-based policy, respectively by introducing the concept of safety layer into the actor network and the concept of PDS into the critic network.

We first characterize the available model concerning the state transition and the corresponding contribution to the reward in the following proposition.
\begin{proposition}
When $\tau \ll \Delta T $, we have
\begin{equation}
\mathrm{Pr} \left(\sum_{i=1}^{N_{\rm s}} \tau p_{ti} = \Delta T \bar p_t\right) = 1~~\text{and}~~ \mathrm{Pr} \left(\sum_{i=1}^{N_{\rm s}} \tau R_{ti} = \Delta T\bar R_t\right) = 1,
\end{equation}
where $\bar p_{t}$ denotes the expectation of transmit power in the $t$th TF over the small-scale fading. Under the optimal power allocation policy, the relationship between $\bar p_{t}$ and ($\alpha_t$, $\bar R_t$)  is
\begin{equation}\label{eqn:ave-pt}
\bar p_t = \bar{p}(\alpha_t, \bar R_t) =   \int_{\frac{\sigma^2}{\alpha_t\xi^{\rm opt}(\bar R_t)}}^{\frac{\sigma^2}{\alpha_t(\xi^{\rm opt}(\bar R_t) - P_{\max})}} \left(\xi^{\rm opt}(\bar R_t) - \frac{\sigma^2}{\alpha_tg_{ti}}\right) \rho(g) {\rm d}g + P_{\max}\int_{\frac{\sigma^2}{\alpha_t(\xi^{\rm opt}(\bar R_t) - P_{\max})}}^{\infty} \rho(g) {\rm d} g.
\end{equation}
 Particularly, for Rayleigh fading and a large value of $P_{\max}$, we have
\begin{equation}\label{eqn:ave-pt-Rt}
\bar p(\bar \alpha_t, R_t) = \frac{\sigma^2}{\alpha_t }\left[ e^{-{\rm E}_1^{-1} \left(\frac{\bar R_t}{W} \ln 2\right)} \left[{\rm E}_1^{-1} \left(\frac{\bar R_t}{W} \ln 2\right)\right]^{-1} - \frac{\bar R_t}{W}\ln 2\right].
\end{equation}
\end{proposition}
\begin{IEEEproof}
By applying the law of large numbers, Proposition 2 can be obtained. Detailed proof is provided in Appendix B.
\end{IEEEproof}

For mobile users in wireless networks, the small-scale channel gains change much faster than large-scale channel gains, hence the condition of $\tau \ll \Delta T$ holds.

\begin{remark}
Proposition 2 shows that the energy to be consumed by the BS in the $t$th TF (i.e.,  $\sum_{i=1}^{N_{\rm s}} \tau p_{ti}$) and the amount of data to be received by the user in the $t$th TF (i.e., $\sum_{i=1}^{N_{\rm s}} \tau R_{ti}$) converge almost surely to their expectations (i.e., the ensemble-average) $\Delta T \bar p_t$ and $\Delta T \bar R_t$, respectively, which can be further computed from \eqref{eqn:ave-pt} or \eqref{eqn:ave-pt-Rt}. 
This means that given an arbitrary action $\bar R_t$,  a part of the state transition (i.e.,  $\sum_{i=1}^{N_{\rm s}} \tau R_{ti}$) and a part of the reward function (i.e., $\sum_{i=1}^{N_{\rm s}} \tau p_{ti}$) can be pre-computed at the beginning of of the $t$th TF, before actually taking the action. This enables us to satisfy the QoS constraint during learning and accelerate learning.
\end{remark}

\subsection{Safety Layer for Actor Network}
In the basic DDPG algorithm of Section III, a penalty term is added to the reward function for ensuring the QoS constraint \eqref{eqn:QoS}. This introduces an extra hyper-parameter $\lambda$, which has to be tuned for striking a tradeoff between the energy minimization against the QoS guarantee. Moreover, such a reward shaping approach can only guarantee constraint satisfaction after the learning converges but cannot guarantee zero constraint violation during the entire learning process due to the random exploration action, which impairs the online performance. To overcome theses issues, we try to meet the QoS constraint by exploiting the model concerning the transitions of the user's buffer state.

According to Proposition 2, by setting the average data rate as $\bar R_t$, the amount of data to be received by the user within the $t$th TF can be pre-computed at the beginning of the TF as $\sum_{i=1}^{N_{\rm s}} \tau R_{ti}  =\Delta T \bar R_t$. Therefore, given the amount of data $\sum_{i=1}^{N_{\rm s}} \tau R_{ti}$ that the user should receive within the $t$th TF for meeting the QoS requirement, the action should be set as
\begin{equation}
\bar{R_t} = \frac{\sum_{i=1}^{N_{\rm s}} \tau R_{ti}}{\Delta T}. \label{eqn:least}
\end{equation}

According to \eqref{eqn:B}, to satisfy the QoS constraint in \eqref{eqn:QoS}, i.e., $B_{t+1} \geq S_{n_{t+1}}$, the least amount of data that should be received by the user within the $t$th TF is $\sum_{i=1}^{N_{\rm s}} \tau R_{ti} \geq\max\{S_{n_{t+1}}- B_t + I(l_t = L_{\rm v})S_{n_t}, 0\}$, which yields
\begin{equation}
\bar R_t \geq \frac{1}{\Delta T}\max\{S_{n_{t+1}}- B_t + I(l_t = L_{\rm v})S_{n_t}, 0\} \label{eqn:QoS2}
\end{equation}
considering \eqref{eqn:least}. To ensure that the executed action is ``safe" in terms of satisfying the constraint $B_{t+1} \geq S_{n_{t+1}}$, we add an additional layer (termed as the \emph{safety layer} \cite{dalal2018safe}) to the output of the original actor network $\mu (\mathbf s_t; \bm \theta_{\mu})$ to adjust the action as follows:
\begin{equation}
{\sf Safety~Layer:~~} a_t =   \max\left\{\mu (\mathbf s_t; \bm \theta_{\mu}) + \mathcal{N}_t, \frac{1}{\Delta T}\max\{S_{n_{t+1}}- B_t + I(l_t = L_{\rm v})S_{n_t}, 0\} \right\}. \label{eqn:SL}
\end{equation}
In this way, the penalty term in \eqref{eqn:reward} can be removed, and hence the hyper-parameter $\lambda$ is no longer needed.

\begin{remark}
Such a safety layer can be also inserted into other DRL algorithms for satisfying the constraints, as long as the constraints can be equivalently transformed into the constraints imposed on the action of each time step with known expressions, as exemplified by \eqref{eqn:QoS2}.\footnote{When the expressions are unknown, some approximation methods can be used~\cite{dalal2018safe}.}

\end{remark}

\subsection{Post-Decision State for Critic Network}
General RL/DRL algorithms are applicable to the scenarios where the dynamic model, including the state transition probability distribution and the reward distribution, are completely unknown. However, for many problems in wireless networks, the dynamic model can be partially known, as exemplified in Proposition 2 for the problem at hand.

To exploit the available partial model for accelerating learning, we introduce PDS to describe the intermediate state between the known dynamic and the unknown dynamic \cite{mastronarde2011fast}. Let $\tilde{\mathbf s}_t$ denote the PDS at the $t$th TF. For the problem considered, $\tilde{\mathbf s}_t$ is defined to characterize the buffer's state, the size of the video segment to be played, as well as the playback progress of the current video segment and the download progress of the whole video after the user receives the transmitted data within the $t$th TF, and to characterize the large-scale channel gains before their transition. To augment our exposition, we write the state vector of the $t$th  and  $(t+1)$th TF together with the defined PDS as follows:
\begin{subequations} \label{eqn:PDS}
	\begin{align}
	{\sf State~at~TF~}t:~&\mathbf s_t  = [B_t, S_{n_t},  l_t,  \eta_t, \bm \alpha_t, \cdots, \bm \alpha_{t-N_t}], \\
	{\sf PDS~at~TF~}t:~&\tilde{\mathbf s}_{t}  \triangleq [B_{t+1}, S_{n_{t+1}}, l_{t+1}, \eta_{t + 1},  \bm \alpha_t, \cdots, \bm \alpha_{t-N_t}], \label{eqn:PDS0}\\
	{\sf State~at~TF~}t+1:~&\mathbf s_{t+1}  = [B_{t+1},  S_{n_{t+1}},  l_{t+1}, \eta_{t + 1}, \bm \alpha_{t+1}, \cdots, \bm \alpha_{t-N_t + 1}].
	\end{align}
\end{subequations}

With the aid of PDS, the state transition can be generally decomposed into a known part ($\mathbf s_t \to \tilde{\mathbf s}_{t}$) and an unknown part ($\tilde{\mathbf s}_t \to \mathbf{s}_{t+1}$). In the following, we first show that benefited from such a decomposition, the high dimensional action-value function, which plays a vital role in many DRL algorithms, can be estimated more efficiently, as a low demensional function, namely the PDS-value function, by analytically deriving the transition probability density function (PDF) accounting for the known transition $\mathbf s_t \to \tilde{\mathbf s}_{t}$. Finally, we use DDPG as an example to show how to integrate PDS with DRL for improving the sample efficiency.

\subsubsection{Estimating the action-value function with PDS}

By introducing the PDS, the reward can be decomposed into two parts formulated as $r_t = r^{\rm k}_{t} + r^{\rm u}_{t}$, where $r^{\rm k}_{t} $ is the reward received from transition $\mathbf{s}_{t} \to \tilde{\mathbf s}_t$ and $r^{\rm u}_{t}$ is the reward received from transition $\tilde{\mathbf s}_t \to \mathbf{s}_{t+1}$.
Let $\rho (\mathbf s_{t+1}, r_t | \mathbf s_t, a_t)$ denote the joint conditional PDF of $\mathbf s_{t+1}$ and $r_t$ when taking action $a_t$ at state $\mathbf s_t$, which characterizes the transition $\mathbf s_t \to \mathbf s_{t+1}$.
If the transition $\tilde{\mathbf s}_t \to \mathbf{s}_{t+1}$ and $r^{\rm u}_t$ are independent from the action $a_t$ (which is true for the problem considered since the transition of large-scale channel gains is independent from $a_t$), we can decompose the joint conditional PDF into known and unknown components as
\begin{equation}
\rho ( \mathbf s_{t+1}, r_t | \mathbf s_t, a_t) = \iint_{(\tilde{\mathbf s}_t, r_{t}^{\rm k})}\rho^{\rm k}\!\left(\tilde{\mathbf s}_t, r^{\rm k}_{t} \big| \mathbf s_t, a_t\right) {\rm d} r^{\rm k}_{t} \rho^{\rm u}\!\left(\mathbf{s}_{t+1}, r_t- r^{\rm k}_{t} \big| \tilde{\mathbf s}_t \right)  {\rm d} \tilde{\mathbf s}_t,
\end{equation}
where the conditional PDF accounting for the transition $\mathbf s_t \to\tilde{\mathbf{s}}_{t}$, i.e., $\rho^{\rm k}(\tilde{\mathbf s}_t, r^{\rm k}_t |\mathbf{s}_{t}, a_t)$, is known (to be derived later), and the conditional PDF accounting for the transition $\tilde{\mathbf s}_t\to \mathbf{s}_{t+1}$, i.e., $\rho^{\rm u} (\mathbf{s}_{t+1}, r_t- r^{\rm k}_{t} | \tilde{\mathbf s}_t )$, is unknown.

Let us define the \emph{PDS-value function} of  $\tilde{\mathbf s}_t$ as the expected accumulated reward achieved by policy $\mu$ started from $\tilde{\mathbf s}_t$, i.e., $V_{\mu}(\tilde{\mathbf s}_t) \triangleq \mathbb E \left[r^{\rm u}_t + \sum_{i=t+1}^{T-1}\gamma^{i-t} r_i \big| \tilde{\mathbf s}_t \right]$. Then, based on the factorization of the state transition by PDS as well as the definitions of the action-value and PDS-value functions, the relationship between the PDS-value function $V_\mu(\cdot)$ and the action-value function $Q_\mu(\cdot)$ can be expressed as
\begin{align}
V_{\mu }(\tilde{\mathbf s}_t) & = \iint_{(\mathbf s_{t+1}, r^{\rm u}_t)}\left[r_t^{\rm u} + \gamma Q_{\mu} (\mathbf s_{t+1}, \mu(\mathbf s_{t+1}))\right] \rho^{\rm u} \!\left(\mathbf{s}_{t+1}, r^{\rm u}_{t} \big| \tilde{\mathbf s}_t \right) {\rm d}r^{\rm u}_{t}{\rm d}\mathbf{s}_{t+1},  \label{eqn:VQ}\\
Q_{\mu}(\mathbf s_t, a_t) & = \iint_{(\tilde{\mathbf s}_t, r^{\rm k}_t)} \left[r^{\rm k}_{t}  +  V_{\mu}(\tilde{\mathbf s}_t) \right] \rho^{\rm k} \!\left(\tilde{\mathbf s}_t, r^{\rm k}_t |\mathbf s_t, a_t\right) {\rm d}r^{\rm k}_t {\rm d}\tilde{\mathbf s}_t. \label{eqn:QV}
\end{align}

By substituting \eqref{eqn:VQ} into \eqref{eqn:QV} and considering $r_t = r^{\rm k}_t + r^{\rm u}_t$ as well as $\rho(\mathbf{s}_{t+1}, r_{t} \big| \mathbf s_t, a_t ) = \iint_{(r^{\rm u}_t, \tilde{\mathbf s}_t)} \rho^{\rm k} (\tilde{\mathbf s}_t, r_t - r^{\rm u}_t |\mathbf s_t, a_t) \rho^{\rm u} (\mathbf{s}_{t+1}, r^{\rm u}_{t} \big| \tilde{\mathbf s}_t ) {\rm d} r^{\rm u}_t {\rm d} \tilde{\mathbf s}_t $, we arrive at
\begin{equation}
Q_{\mu}(\mathbf s_t, a_t) = \iint_{(\mathbf s_{t+1}, r_t)}\left[r_t + \gamma Q_{\mu} (\mathbf s_{t+1}, \mu(\mathbf s_{t+1}))\right] \rho\left(\mathbf{s}_{t+1}, r_{t} \big| \mathbf s_t, a_t \right) {\rm d}r_{t}{\rm d}\mathbf{s}_{t+1}, \label{eqn:bellman}
\end{equation}
which is actually the Bellman equation with respect to $Q_{\mu}$, based on which the critic network parameter $\bm \theta_{Q}$ is updated by the DDPG. 

Considering that \eqref{eqn:bellman} can be derived from \eqref{eqn:VQ} and \eqref{eqn:QV}, we can directly develop corresponding RL algorithm based on \eqref{eqn:VQ} and \eqref{eqn:QV} rather than \eqref{eqn:bellman}.
Furthermore, since the transition PDF $\rho^{\rm k} \!\left(\tilde{\mathbf s}_t, r^{\rm k}_t |\mathbf s_t, a_t\right)$ can be derived, the action-value function $Q_{\mu}(\mathbf s_t, a_t)$ can be obtained by estimating the PDS-value function $V_{\mu}(\tilde{\mathbf s}_t)$ in the right-hand-side (RHS) of \eqref{eqn:QV}. 

\begin{remark}
Compared to directly estimating $Q_{\mu}(\mathbf s_t, a_t)$, estimating $V_{\mu}(\tilde{\mathbf s}_t)$ can be more sample-efficient, since  $V_{\mu}(\tilde{\mathbf s}_t)$  no longer depends on the action.
\end{remark}

\subsubsection{Deriving the transition PDF of $\mathbf s_t \to\tilde{\mathbf{s}}_{t}$}
In what follows, we derive $\rho^{\rm k} (\tilde{\mathbf s}_t, r^{\rm k}_t |\mathbf s_t, a_t)$ for the considered video streaming problem. In particular, we show that $ \tilde{\mathbf s}_t$ and $r^{\rm k}_t$ become  deterministic given that the agent executes action $a_t$ at state  $\mathbf s_t$. 

According to Proposition 2, we have $\sum_{i=1}^{N_{\rm s}}\tau R_{ti} = \Delta T \bar R_t = \Delta T a_t$. Upon substituting this into \eqref{eqn:B}, the first element of $\tilde{\mathbf s}_t$, i.e., the buffer state $B_{t+1}$, can be expressed as
\begin{equation}
\tilde{\mathbf s}_t[1] = B_{t+1} = B_t + \Delta T a_t - I(l_t = L_{\rm v})S_{n_t} = \mathbf s_{t}[1] + \Delta T a_t - I(\mathbf{s}_{t}[3] = L_{\rm v})\mathbf{s}_{t}[2], \label{eqn:s1}
\end{equation}
which is deterministic, given $\mathbf s_t$ and $a_t$. The second element of $\tilde{\mathbf s}_t$ is $\tilde{\mathbf s}_t[2] = S_{n_{t+1}}$, i.e., the size of the video segment to be played at the $(t+1)$th TF,  which is also deterministic given $\mathbf s_t$, because the size of each segment is known after the user issues the video request. The third element is $\tilde{\mathbf s}_t[3] = l_{t+1}$, i.e., the playback progress of the video segment to be played at the $(t+1)$th TF. By substituting \eqref{eqn:s1} into  \eqref{eqn:lt}, $l_{t+1}$ can be expressed as a deterministic function of $\mathbf s_t$ and~$a_t$ as
\begin{equation}
f_{\rm l}(\mathbf s_t, a_t) =\left\{\begin{array}{ll}
{\rm mod}(l_t, L_{\rm v}), &~\text{if}~S_{n_{t+1}} > \mathbf s_{t}[1] + \Delta T a_t - I(\mathbf{s}_{t}[3] = L_{\rm v})\mathbf{s}_{t}[2],  \\
{\rm mod}(l_t, L_{\rm v}) + 1, &~\text{otherwise.}  \\
\end{array}
\right.
\end{equation}
The fourth element is $\tilde{\mathbf s}_t[4] = \eta_{t + 1}$, i.e., the download progress of the whole video file. Based on \eqref{eqn:download} and $\sum_{i=1}^{N_{\rm s}}\tau R_{ti} = \Delta T a_t$ , we can obtain
\begin{equation}
\eta_{t+1} = \frac{\eta_t + \sum_{i=1}^{N_{\rm s}}\tau R_{ti}}{\sum_{n=1}^{N_{\rm v}} S_n}= \frac{\eta_t + \Delta T a_t}{\sum_{n=1}^{N_{\rm v}} S_n}.
\end{equation}
The rest of the elements in $\tilde{\mathbf s}_t$ are the same as those in $\mathbf s_t$. Finally, $\tilde{\mathbf s}_t$ can be  expressed as a function of $\mathbf s_t$ as
\begin{align}
\tilde{\mathbf s}_t = \mathbf{f}_{\rm PDS} (\mathbf s_t, a_t) =  \left[\mathbf s_{t}[1] + \Delta T a_t - I(\mathbf{s}_{t}[3] = L_{\rm v})\mathbf{s}_{t}[2], S_{n_{t+1}}, f_{\rm l}(\mathbf{s}_t, a_t), \tfrac{\eta_t + \Delta T a_t}{\sum_{n=1}^{N_{\rm v}} S_n},  \mathbf{s}_t [5\!:]\right], \label{eqn:fpds}
\end{align}
where $\mathbf s_t [x\!:]$ denotes the sliced vector containing from the $x$th element to the last element of $\mathbf s_{t}$.

Again, according to the definition of $\tilde{\mathbf s}_t$, we can obtain $r^{\rm k}_{t} = r_t$ and $r^{\rm u}_t = 0$. Furthermore, considering Proposition 2, we have
\begin{equation}
\left\{\begin{array}{ll}r^{\rm k}_{t} & = r_t = \sum_{i=1}^{N_{\rm s}} \tau p_{ti} = \Delta T \bar p (  \alpha_t, \bar R_t) \triangleq \Delta T \bar p (\mathbf s_t,  a_t), \\
r^{\rm u}_t &= 0,
\end{array}\right. \label{eqn:r}
\end{equation}
where both $r^{\rm k}_{t}$ and $r^{\rm u}_t$ are deterministic given $\mathbf{s}_t$ and $a_t$, and we rewrite $\bar p (\alpha_t,  \bar R_t)$ into $\bar p (\mathbf s_t, a_t)$ because $\alpha_t$ is an element of $\mathbf s_t$ and $a_t = \bar R_t$. Thus, the transition PDF $\rho^{\rm k} \!\left(\tilde{\mathbf s}_t, r^{\rm k}_t \big|\mathbf s_t, a_t\right)$ can be expressed as
\begin{equation}
\rho^{\rm k} \left(\tilde{\mathbf s}_t, r^{\rm k}_t \big| \mathbf{s}_t, a_t  \right) = \delta\left(\tilde{\mathbf s}_t - \mathbf{f}_{\rm PDS} (\mathbf s_t, a_t) , r^{\rm k}_t - \Delta T \bar p ( \mathbf s_t, a_t )\right),   \label{eqn:rhok}
\end{equation}
where $\delta(\tilde{\mathbf s}_t -x, r^{\rm k}_t- y)$ denotes the two-dimensional Dirac delta function defined as $\int_{\tilde{\mathbf s}_t} \int_{r^{\rm k}_t} \delta(\tilde{\mathbf s}_t -x , r^{\rm k}_t -y ) {\rm d} \tilde{\mathbf s}_t {\rm d} r^{\rm k}_t = 1$, and $\delta(\tilde{\mathbf s}_t -  x, r^{\rm k}_t - y) = 0$ if $ \tilde{\mathbf s}_t   \neq x$ or $r^{\rm k}_t \neq y$.

Finally, by substituting \eqref{eqn:r} and  \eqref{eqn:rhok} into \eqref{eqn:VQ} and  \eqref{eqn:QV}, we can obtain the relationship between $V_\mu(\cdot)$ and $Q_\mu(\cdot)$ for our video streaming problem as
\begin{align}
V_{\mu}(\tilde{\mathbf s}_t) & =  \gamma\int_{\mathbf s_{t+1}} Q_{\mu} (\mathbf s_{t+1}, \mu(\mathbf s_{t+1})) \rho^{\rm u}\left(\mathbf{s}_{t+1} \big| \tilde{\mathbf s}_t \right){\rm d}\mathbf{s}_{t+1}, \label{eqn:VQ2}\\
Q_{\mu}(\mathbf s_t, a_t) & = \Delta T \bar p \left( \mathbf s_t, a_t \right) + V_{\mu}\left( \mathbf{f}_{\rm PDS} (\mathbf s_t, a_t) \right). \label{eqn:QV2}
\end{align}

\begin{figure}[!htb]
	\vspace{-5mm}
	\centering
	\includegraphics[width=0.85\textwidth]{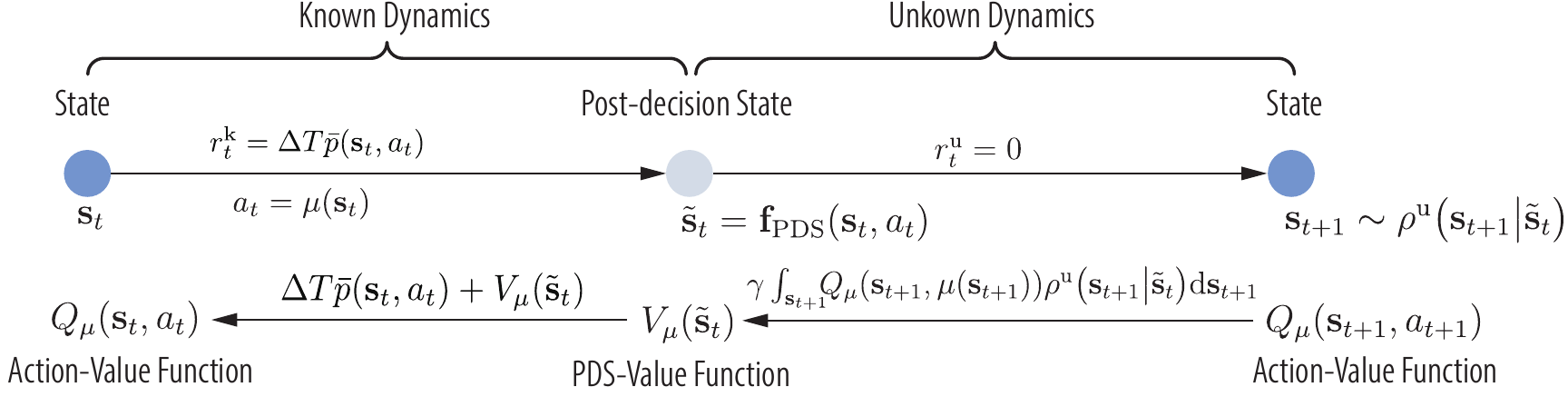}
	\vspace{-2mm}
	\caption{The relations between states $\mathbf s_{t}$, $\mathbf s_{t+1}$ and the PDS $\tilde{\mathbf s}_{t}$.}
	\label{fig:PDS}
\end{figure}

In Fig.~\ref{fig:PDS}, we summarize the relationship between $\mathbf s_t$, $\tilde{\mathbf s}_t$ and $\mathbf{s}_{t+1}$.

\subsubsection{PDS-DDPG algorithm}
\begin{figure}[!htb]
	\vspace{-1mm}
	\centering	
	\subfigure[Modified actor network, $\mu_{\rm s}(\mathbf s_t, \mathcal{N}_t; \bm \theta_{\mu})$]{
		\label{fig:pdsddpg-a} 
		\includegraphics[height=0.22\textwidth]{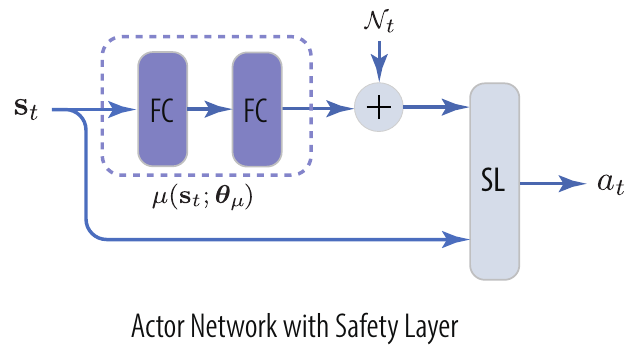}}
	\subfigure[Modified critic network, $Q(\mathbf s_t, a_t; \bm \theta_{V})$]{
		\label{fig:pdsddpg-c} 
		\includegraphics[height=0.22\textwidth]{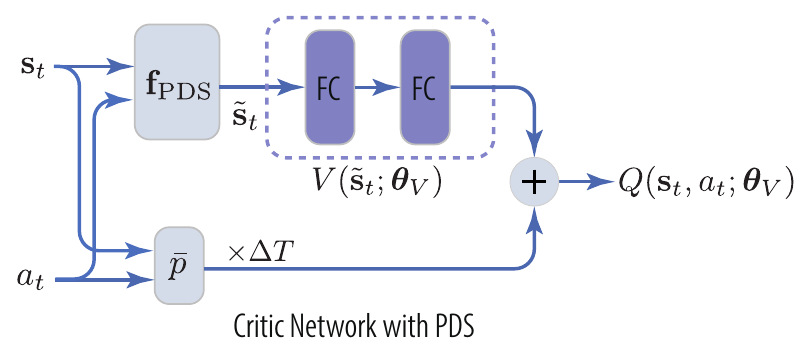}}
	\caption{Architecture of the actor and critic networks for enhanced DDPG with PDS and safety layer (SL).}
	\label{fig:pdsddpg}
	\vspace{-3mm}
\end{figure}

In the following, we use DDPG as an example to show how to integrate PDS with DRL algorithms. Again, we parameterize the transmission policy $\mu$ by a NN with the parameter $\bm \theta_{\mu}$. Upon adding the safety layer defined by \eqref{eqn:SL}, the structure of the modified actor network $\mu_{\rm s}(\mathbf s_t, \mathcal{N}_t; \bm \theta_{\mu})$ is shown in Fig~\ref{fig:pdsddpg-a}. Based on \eqref{eqn:QV2}, we use a NN $V(\tilde{\mathbf s}_t;\bm \theta_{V})$ to approximate $V_{\mu}(\tilde{\mathbf s}_t)$ and then obtain  the approximated $Q_{\mu }(\mathbf s_t, a_t)$ as
\begin{equation}
 Q(\mathbf s_t, a_t; \bm \theta_V)  = \Delta T \bar p \left(\mathbf s_t, a_t \right) + V\left( \mathbf{f}_{\rm PDS} (\mathbf s_t, a_t);\bm \theta_{V} \right),  \label{eqn:Qv}
\end{equation}
whose structure is shown in Fig.~\ref{fig:pdsddpg-c}.

\begin{remark}
By comparing Fig.~\ref{fig:pdsddpg-c} and Fig.~\ref{fig:ddpg-c}, we can see that the partially known dynamic model has been integrated into the modified critic network, where $\mathbf{f}_{\rm PDS}(\mathbf s_t, a_t)$ and $\Delta T \bar p (\mathbf s_t, a_t )$ characterize the known part of the state transition and the reward function, respectively. Consequently, less information has to be learned, and hence the number of unknown parameters in the modified critic network can be reduced, which accelerates the learning. 
\end{remark}

Analogous to the update rule of $\bm \theta_{Q}$ in \eqref{eqn:Q} based on the Bellman equation \eqref{eqn:bellman}, the update rule of $\bm \theta_V $ for the modified critic network can be obtained based on \eqref{eqn:VQ2} as
\begin{equation}
\bm \theta_{V} \leftarrow \bm \theta_{V} - \frac{\delta_{V}}{|\mathcal{B}|}\nabla_{\bm \theta_{V}}\sum_{j\in \mathcal{B}}\left[y_j - V(\mathbf{f}_{\rm PDS} (\mathbf s_j, a_j);\bm \theta_{V})\right]^2, \label{eqn:Q2}
\end{equation}
where we have substituted $\tilde{\mathbf s}_t = \mathbf{f}_{\rm PDS} (\mathbf s_t, a_t)$, $y_j = 0$ if all the segments have been transmitted to the user, and $y_j =  \gamma Q'(\mathbf s_{j+1}, \mu'_{\rm s}(\mathbf s_{j+1}, 0;\bm \theta'_{\mu}); \bm \theta_{V}') $ otherwise. Again, $Q'(\cdot; \bm\theta_{V}')$ and $\mu'_{\rm s}(\cdot;\bm  \theta_{\mu}')$ are the target critic network and target actor network, respectively, which have the same structure as $Q(\cdot; \bm\theta_{V})$ and $\mu_{\rm s}(\cdot; \bm\theta_{\mu} )$, and are respectively updated by $\bm \theta_V' \leftarrow  \omega \bm\theta_V +  (1 - \omega) \bm \theta_{V}'$ and $\bm \theta_\mu' \leftarrow  \omega \bm \theta_\mu +  (1 - \omega) \bm \theta_\mu'$.

From \eqref{eqn:Qv}, we can arrive at
\begin{equation}
\nabla_{a} Q(\mathbf s_t, a;\bm \theta_{V}) =  \Delta T\nabla_{a}\bar p(\mathbf s_t, a) + \nabla_{\tilde{\mathbf s}} V(\tilde{\mathbf s};\bm \theta_{V})\big|_{\tilde{\mathbf s} = \mathbf{f}_{\rm PDS} (\mathbf s_t, a)} \nabla_{a}\mathbf{f}_{\rm PDS} (\mathbf s_t, a), \label{eqn:dQv}
\end{equation}
By substituting \eqref{eqn:dQv} into \eqref{eqn:mu}, we can derive the update rule of $\bm \theta_{V}$ for the modified actor network as
\begin{align}
\bm \theta_{\mu} \leftarrow \bm \theta_{\mu} +  \frac{\delta_\mu}{|\mathcal{B}|} \sum_{j\in\mathcal{B}} \Big[&\Delta T\nabla_{a}\bar p(\mathbf s_t, a)+ \nonumber \\
& \nabla_{\tilde{\mathbf s}} V(\tilde{\mathbf s};\bm \theta_{V}) \nabla_{a}\mathbf{f}_{\rm PDS} (\mathbf s_j, a)\Big]\Big|_{a = \mu_{\rm s}(\mathbf s_j,0; \bm \theta_{\mu}),~ \tilde{\mathbf s} = \mathbf{f}_{\rm PDS} (\mathbf s_j, a)} \nabla_{\bm \theta_\mu} \mu_{\rm s} (\mathbf s_j, 0; \bm \theta_{\mu}). \label{eqn:mu2}
\end{align}

\begin{remark}
	\normalfont 
	The relationship between the PDS-value function and the action-value function, i.e., \eqref{eqn:VQ} and \eqref{eqn:QV}, are also applicable to other MDP tasks as long as: 1) Parts of the state transition are known; 2) The transition from PDS $\tilde{\mathbf s}_t$ to the next state $\mathbf{s}_{t+1}$ is independent from  $a_{t}$. Therefore, the proposed approach of incorporating PDS into the DRL algorithm can be extended to other wireless tasks. Apart from DDPG, this approach can be also implemented upon other DRL algorithms that involves the estimation of action-value function, e.g., TD3~\cite{fujimoto2018addressing} and Q-Prop~\cite{gu2016q}.
\end{remark}

\subsection{Virtual Experiences}
 The analysis in the previous subsection suggests that the state transition $\mathbf s_t[1\!:\!4] \to \mathbf s_{t+1}[1\!:\!4]$ and the reward can be obtained in advance, given $\mathbf s_t$ and $a_t$. Further considering that the transition of the average channel gain does not depend on $\mathbf s_t[1\!:\!4]$ and $a_t$, we are able to generate virtual experiences based on historical traces of average channel gains recorded for previously served users. The virtual experiences can then be used for training the NNs for further accelerating the learning procedure by relying on less interactions with environment.

Specifically, let $\bm h^{(j)} = [\bm \alpha_t^{(j)}]_{t=1-N_t, \cdots, T}$ denote a trace of the average channel gains of a previously served user recorded during the $j$th video streaming episode. From $\bm h^{(j)}$ we can generate an initial state for a virtual user  as $\mathbf s_1 = [B_1, S_{n_1}, l_1, \eta_1, \bm \alpha_1^{(j)}, \cdots, \bm \alpha_{1-N_t}^{(j)}]$ and obtain the action output by the actor network as $a_1 = \mu_{\rm s}(\mathbf s_1, \mathcal{N}_1; \bm \theta_{\mu})$. Let the average channel gains of the virtual user evolve the same as recorded in the historical trace $\bm h^{(j)}$, then the next state and the reward can be directly computed as $\mathbf s_2 = [\mathbf{f}_{\rm PDS} (\mathbf s_1, a_1)[1\! : \!4], \bm \alpha_{2}^{(j)}, \cdots, \bm \alpha_{2-N_t}^{(j)}]$ and  $r_1 = \Delta T \bar p \left( a_1 \right)$ based on \eqref{eqn:fpds} and  \eqref{eqn:r}, respectively. This suggests that the agent can deduce how the episode continues for a given transmission policy and channel trace. Hence, it can generate virtual experiences accordingly, given that $\bm h^{(j)}$ is a true channel trace sampled from the same wireless environment, the virtual experiences can be used for training both the actor and the critic networks.

To generate and exploit the virtual experiences, every time a real episode terminates, we store the channel trace into the channel trace buffer $\mathcal{H}$ and randomly sample $K$ traces from $\mathcal{H}$ to generate $K$ virtual episodes. The virtual experiences are stored into the experience relay buffer $\mathcal{D}$ so that the virtual experiences can be sampled together with the real experiences for training both the actor and critic networks. The whole learning procedure of PDS-DDPG using virtual experiences is shown in Algorithm \ref{alg2}.

\begin{algorithm}[!htb]
	\linespread{1}
	\caption{\small Transmission Policy Learning Based on PDS-DDPG with Virtual Experience}\
	\label{alg2}
	\small
	\begin{algorithmic}[1]
		\State Initialize modified critic and actor networks $Q(\mathbf s, a;\bm \theta_V)$ and $\mu_{\rm s} (\mathbf s, \mathcal{N}; \bm \theta_{\mu})$ with random weights $\bm \theta_{V}$, $\bm \theta_{\mu}$.
		\State Initialize target networks $Q'$ and $\mu'$ with weights $\bm \theta_{V}'	\leftarrow \bm \theta_{V}$, $\bm \theta_{\mu}' \leftarrow \bm \theta_{\mu}$.
		\For{${\tt episode} =1,2, \cdots $}
			\State Observe initial state $\mathbf s_1$ from the environment, and initialize channel trace $\bm h \leftarrow [\bm \alpha_{1-N_t}, \cdots, \bm \alpha_1]$.
			\For{TF $t = 1, 2, \cdots $} \Comment{$\tt Real~episode$}
				\State Select action $a_t = \mu(\mathbf s_t, \mathcal{N}_t; \bm \theta_{\mu}) $ with exploration noise, and set $\bar {R}_t = a_t$ and $\xi_t = \xi^{\rm opt}(\bar R_t)$.
				\For{TS $i = 1, \cdots, N_{\rm s}$}
					\State   Allocate transmit power according to \eqref{eqn:popt}.
				\EndFor
				\State Observe reward $r_t$, new state $\mathbf s_{t+1}$, and add $\bm \alpha_{t+1}$ into channel trace $\bm h \leftarrow [\bm h, \bm \alpha_{t+1}]$.
	
				\State Store $[\mathbf{s}_t, a_t, \mathbf s_{t+1}]$  in $\mathcal D$.	
					\State Randomly sample a mini-batch from~$\mathcal{D}$, update the actor and critic networks according to \eqref{eqn:mu2} and~\eqref{eqn:Q2}.
					\State Update the target networks $\bm \theta_{\mu}' \leftarrow \omega {\bm\theta}_{\mu} + (1-\omega) {\bm \theta}_{\mu}'$, ${\bm \theta}_{V}' \leftarrow \omega {\bm\theta}_{V} + (1-\omega) {\bm \theta}_{V}'$	
				\If{all video segments have been transmitted to the user}
					\State Store channel trace $\bm h$ in $\mathcal{H}$, set $\tt done = 0$ and  {\bf break}
				\EndIf
			\EndFor	
			\For{ $k = 1, \cdots, K$} 
				\State Randomly sample a channel trace $\bm h^{(j)}$ from $\mathcal{H}$ and generate initial state $\mathbf s_1$.
				\For{$t = 1, 2, \cdots $} \Comment{$\tt Virtual~episode$}
					\State Select action $a_t = \mu(\mathbf s_t, \mathcal{N}_t; \bm \theta_{\mu})$ with exploration noise.
					\State Obtain $\bm \alpha_{t+1}$ from $\bm h^{(j)}$, and obtain next state $\mathbf s_{t+1} = [\mathbf{f}_{\rm PDS} (\mathbf s_t, a_t)[1\!:\!4], \bm \alpha_{t+1}^{(j)}, \cdots, \bm \alpha_{t-N_t + 1}^{(j)}]$
					\State Store $[\mathbf{s}_t, a_t, \mathbf s_{t+1}]$  in $\mathcal D$ and repeat steps 11$\sim$12.	
					\If{all video segments have been transmitted to the virtual user}
						\State {\bf break}
					\EndIf
				\EndFor
			\EndFor
		\EndFor
	\end{algorithmic}
\end{algorithm}

\vspace{-7mm}
\begin{remark}
The way we generate virtual experiences can be extended to other RL problems, as long as the unknown dynamics are independent from the known dynamics. This is true for many problems in wireless networks, where the dynamics of wireless channel do not depend on the action and the transition of other elements in the state.
\end{remark}

\section{Simulation Results}
In this section, we evaluate the performance of the proposed DRL-based policies by comparing them to the benchmark policies via simulations.

\vspace{-3mm}
\subsection{Simulation Setup}
\begin{figure}[!htb]
	\vspace{-3mm}
	\centering	
	\subfigure[One road]{
		\label{fig:layout1} 
		\includegraphics[width=0.315\textwidth]{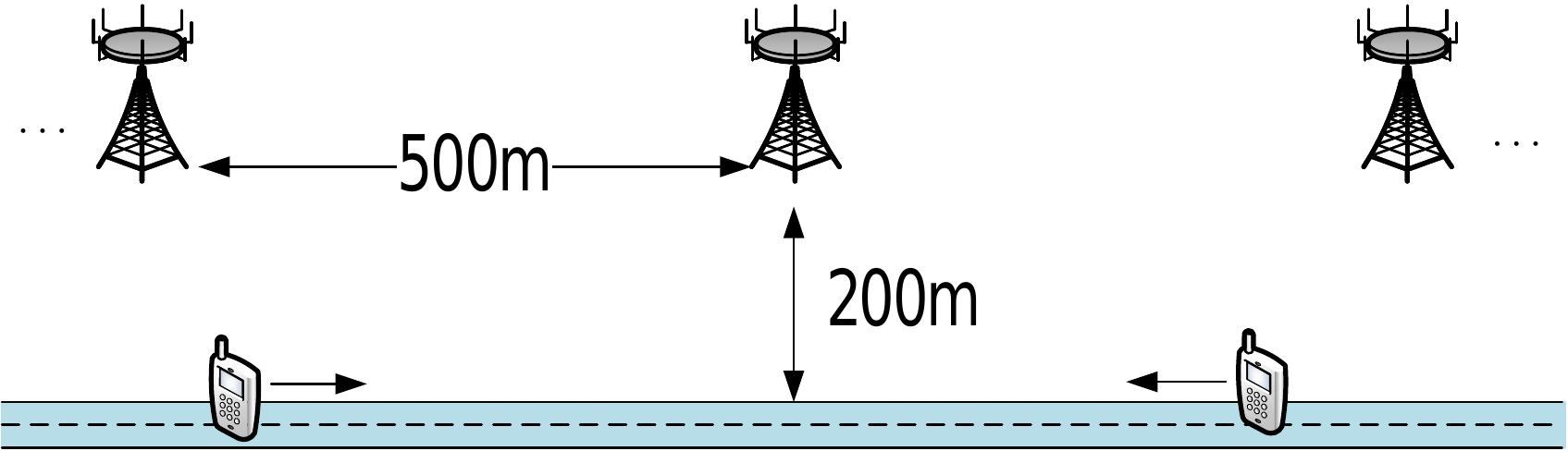}}
	\subfigure[Multiple roads]{
		\label{fig:layout2} 
		\includegraphics[width=0.315\textwidth]{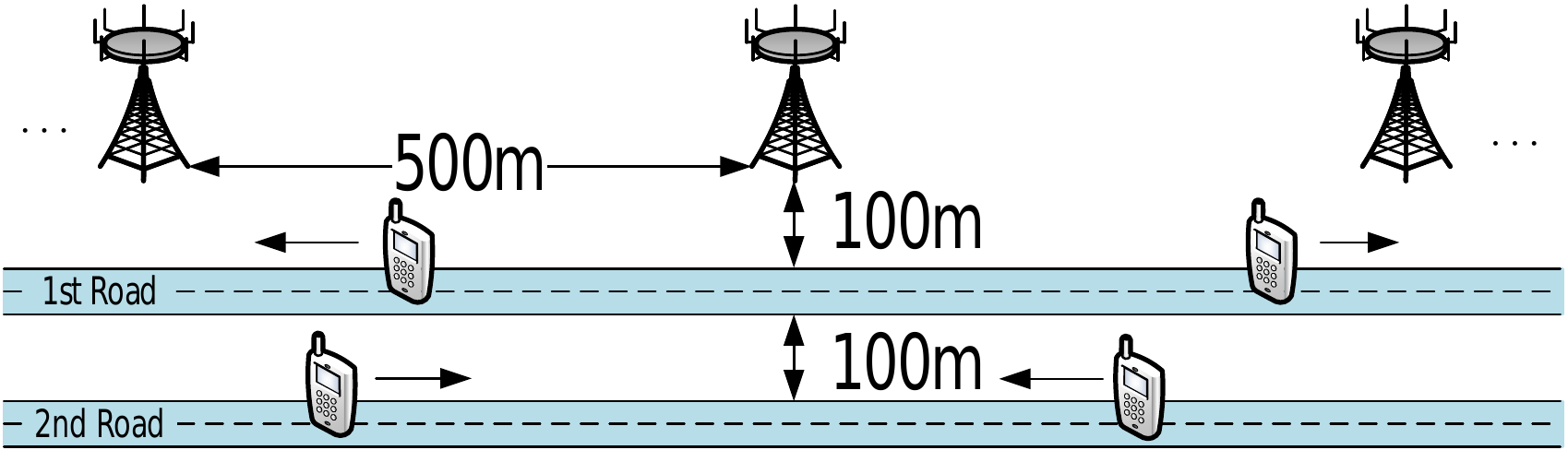}}
	\subfigure[Traffic light]{
		\label{fig:layout3} 
		\includegraphics[width=0.315\textwidth]{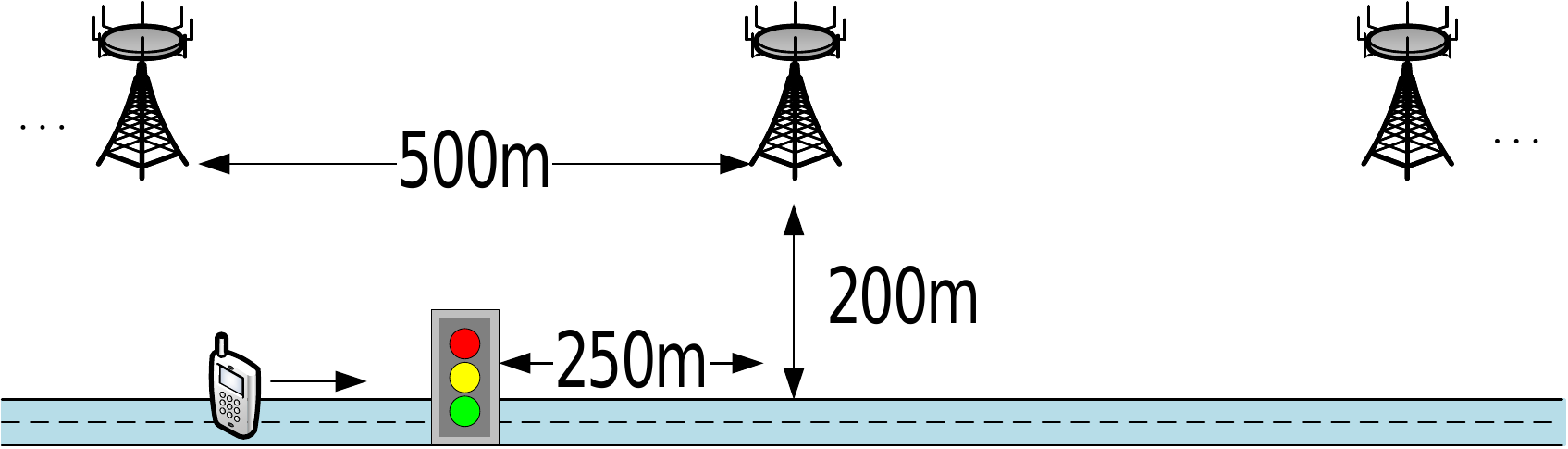}}
	\caption{Simulation scenarios.}
	\label{fig:layout}
\end{figure}
We consider several scenarios, where users are moving along one or multiple roads across multiple cells, as shown in Fig. \ref{fig:layout}. The distance between the adjacent BSs is $500$ m and the maximal transmit power of each BS is $46$ dBm. The noise power is $-95$ dBm and the transmission bandwidth of each user is $20$ MHz. Since the circuit power consumption is the same for all the  policies considered, we only evaluate the transmit energy consumed by video streaming. The path loss is modeled as $35.3+37.6\log_{10} (d)$ in dB where $d$ is the distance between the user and BS in meters. The small-scale channel is Rayleigh fading. The playback duration of each video file is $150$ s and that of each segment is $10$ s. In practice, variable bitrate encoding may be used to assign higher (lower) video bitrate for more (less) visually complex video segment. To reflect general VBR videos without targeting a specific video file, the video bitrate of each segment is generated from Gaussian distribution with an average value of $8$ Mbps~\cite{youtube} and with a standard deviation of $0.3$ Mbps. The duration of each TF is $\Delta T = 1$ s and the duration of each TS is $\tau = 1$ ms.

\subsection{Tuned Parameters in Algorithm 1 and Algorithm 2}
After fine-tuning, the NNs in our algorithms are configured as follows:
\begin{enumerate}
	\item Algorithm 1: For the actor network, $\mu(\cdot)$ has two fully-connected layers each with $200$ nodes.  For the critic network, the state and action are first concatenated and then go through two fully-connected layers each having $200$ nodes.
	\item Algorithm 2: By employing PDS and the safety layer, the number of nodes in each hidden layer of the modified actor and critic networks is reduced to $100$ nodes. Consequently, the total number of unknown parameters to be learned in the modified networks is reduced roughly by a factor of four compared with the original actor and critic networks of Algorithm~1. This enables faster convergence speed and lower computational complexity.
\end{enumerate}

 All the hidden layers of the above NNs use the rectified linear units (ReLU) as the activation function. The output layer of the critic and modified critic networks has no activation function. The network $\mu(\cdot)$ of both the actor and the modified actor networks employs $40\times [\tanh(x) + 1]$ as the activation function, which bounds the transmission data rate within $[0, 80]$ Mbps. For the state representation, we set $N_{\rm b} = 2$ and $N_{\rm t} = 2$. 

The learning rate is $\delta_\mu = 10^{-4}$, $\delta_Q = \delta_V = 10^{-3}$ and the update rate for the target  networks is $\omega = 10^{-3}$~\cite{DDPG}. To determine the most appropriate  mini-batch size for gradient descent, we have tried the parameter in $\{64, 128, 256, 512, 1024\}$ and found that $|\mathcal B| = 1024$ performs the best. The discount factor is set to $\gamma= 1$ because we aim to minimize the total energy consumption of each video session. The noise term $\mathcal{N}_t$ obeys the Gaussian distribution with zero mean and a standard deviation reduced linearly from $10$ to zero during the training phase and remains zero during the testing phase. All the testing results of PDS-DDPG are trained for $1000$ episodes and all the empirical cumulative distribution function (CDF) curves are obtained from 1000 testing episodes with different random seeds. More details of the experiments can be found in Appendix C.

\vspace{-3mm}
\subsection{Performance Evaluation}
We compare the proposed DRL-based polices to the following baselines:
\begin{itemize}
	\item \emph{Non-predictive}: This is the existing power allocation policy designed without using any future information. To satisfy the QoS constraint, the BS maintains an constant average data rate within each TF as $\bar R_t = S_{n_{t+1}}/L_{\rm v}$ so that the segment to be played in the next TF is downloaded in the current TF. Without using any predicted information, this policy can only minimize the average energy consumption within the current TF via the power allocation policy given by \eqref{eqn:popt}.
	\item \emph{Optimal}: This method assumes perfect values of future average channel gains and optimizes power allocation using the algorithm proposed in~\cite{scy}.
	\item \emph{LSTM \& Optimize}: This method employs a long short-term memory network (LSTM)  to predict the future average channel gains within a prediction window and then optimizes  power allocation using~\cite{scy}, which reflects the practical performance of first-predict-then-optimize PRA policy in the presence of prediction errors. A too long prediction window  (say $150$ s, the same as the duration of video playback) incurs large prediction errors, which degrade the gain of PRA. One the contrary, a too short prediction window will not provide enough future information to reap the gain of PRA. To strike a trade-off, we have tried the prediction window length in $\{30, 60, 90, 120\}$ s and found $60$ s is the best. More details about the implementation of this method can be found in Appendix C.
\end{itemize}

In what follows, we compare the performance of the policies in different scenarios.

\subsubsection{Same Road \& Constant Speed}
We first consider the scenario shown in Fig.~\ref{fig:layout1}, where each user moves along the same road at a constant velocity of $15$ m/s. In this case, the future average channel gains can be perfectly predicted. Therefore, we can use the optimal policy to obtain the energy consumption lower bound.

\begin{figure}[!htb]
	\vspace{-3mm}
	\centering	
	\subfigure[Energy consumption per episode. The lines reflect the average value and the shaded bands reflects the standard deviation.]{
		\label{fig:learning} 
		\includegraphics[width=0.487\textwidth]{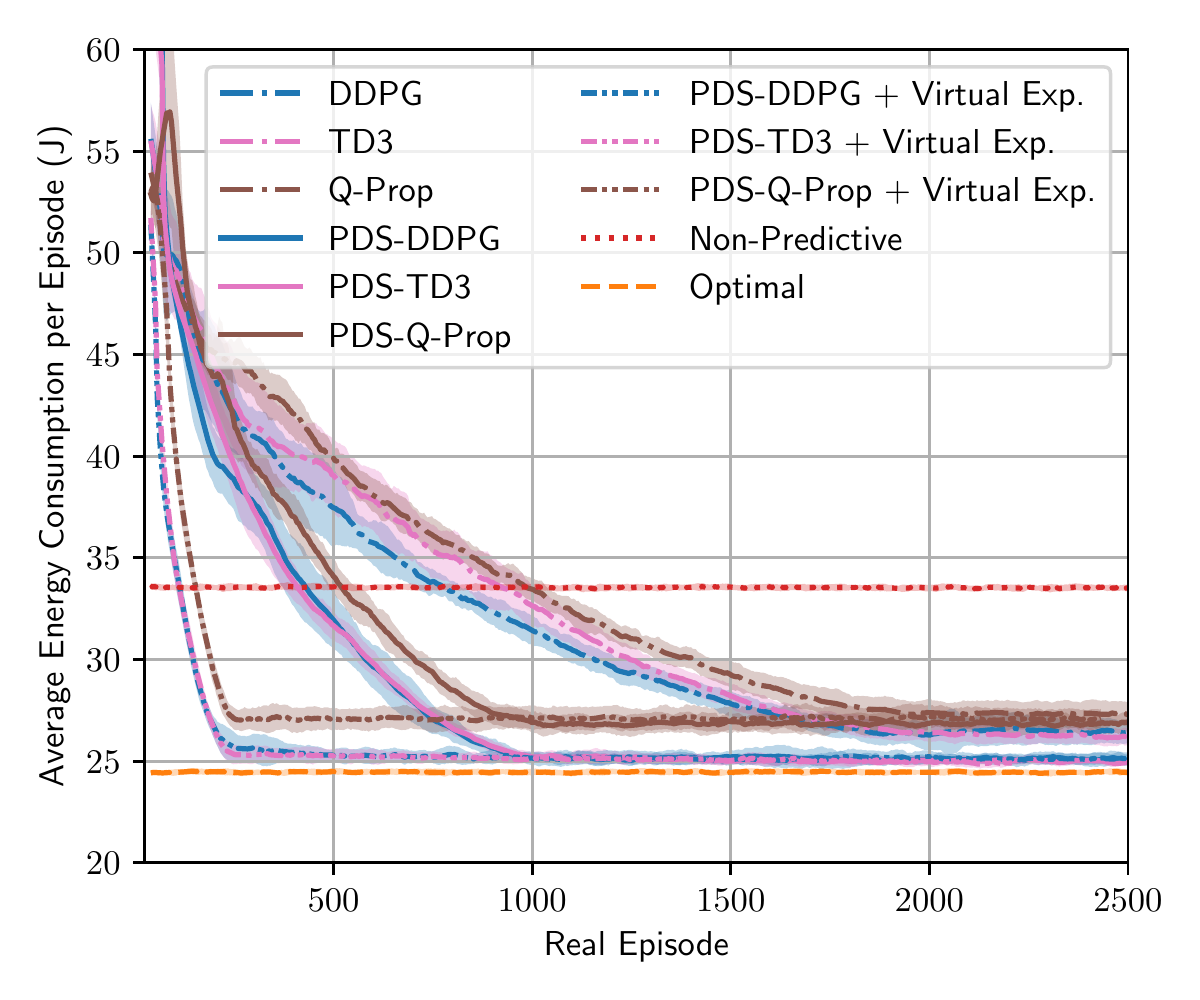}}
	\subfigure[ QoS constraint violation.]{
		\label{fig:constraint} 
		\includegraphics[width=0.487\textwidth]{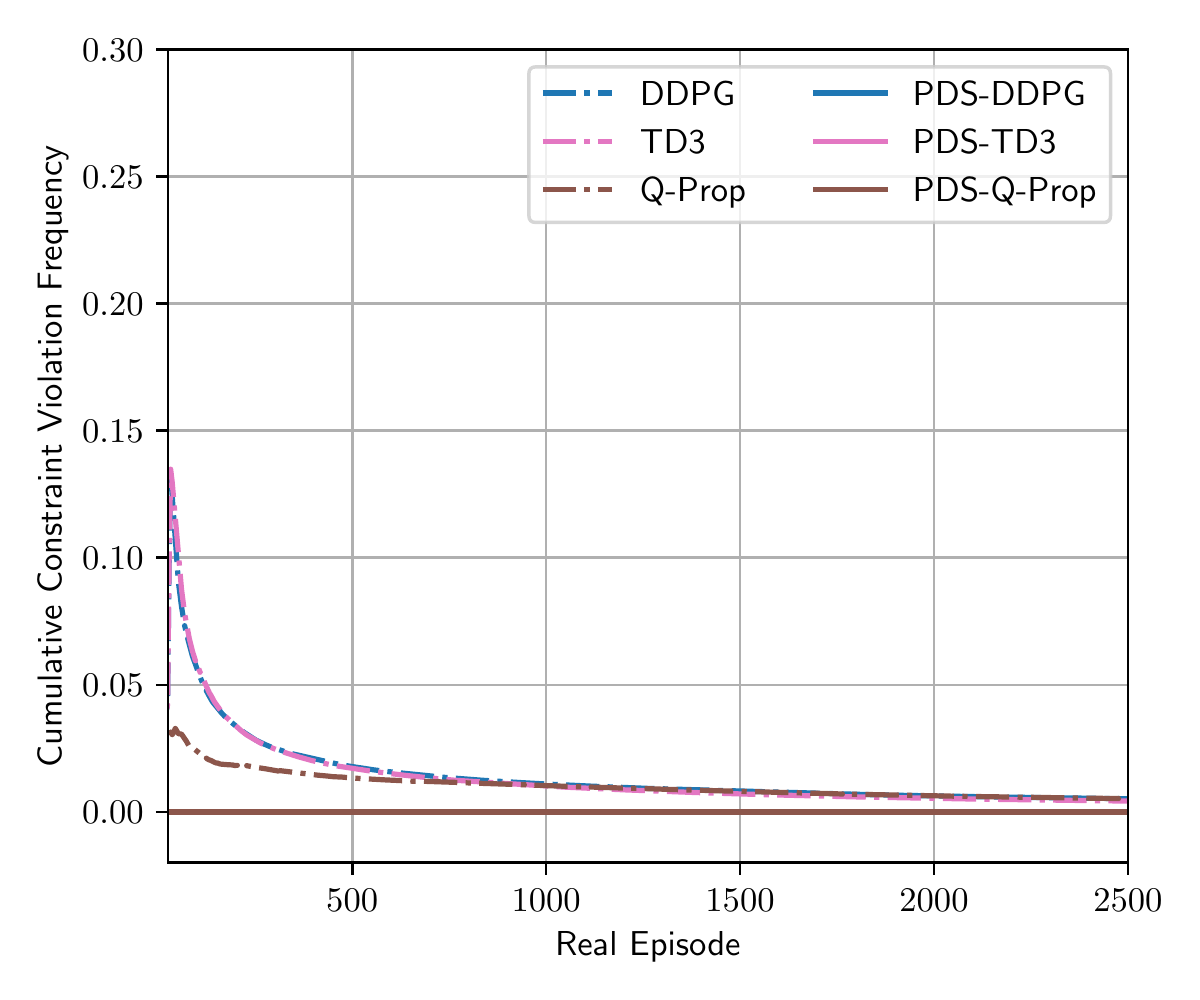}}
	\caption{Learning curves. All experiments are run for $20$ different random seeds each. The curves are smoothed by averaging over a window of $40$ episodes. The penalty coefficient in the reward function is set as $\lambda = 30$ for the algorithms without safety layer and the virtual episode frequency in Algorithm 2 is set to $K=4$ for ``PDS-DDPG + Virtual Exp". Detailed settings for (PDS-)TD3 and (PDS-)Q-Prop are provided in Appendix C.}
	\label{fig:learning constraint}
	\vspace{-2mm}
\end{figure}

In Fig. \ref{fig:learning}, we show the learning curves of the proposed DRL-based policies. To show that our proposed DRL framework is not limited to DDPG, we also implement two other state-of-the-art DRL algorithms, namely TD3~\cite{fujimoto2018addressing} and Q-Prop~\cite{gu2016q}. Considering that the major concern of RL convergence speed lies in the number of interactions between the agent and the environment, the $x$-axis represents the number of real episodes. We can see that the proposed DRL-based policies converge close to the optimal policy and they outperform  ``\emph{Non-predictive}" after convergence. By exploiting the partial model about the system dynamics, all PDS-based algorithms outperform their non-PDS counterparts both in terms of the convergence speed and average energy consumption. For example, by employing PDS and the safety layer, PDS-DDPG converges twice faster than DDPG. By further training using both the real and virtual experiences, all PDS-based algorithms with ``\emph{Virtual Exp.}"  converge $10$ times faster than their non-PDS counterparts.  In Fig.~\ref{fig:constraint}, we show the constraint violation during the learning process, where the \emph{cumulative constraint violation frequency} is the ratio of cumulative number of TFs with constraint violation to the total number of TFs. By adopting the safety layer, all PDS-based algorithms can satisfy the QoS constraint during the entire learning process.
Considering that the performances of PDS-DDPG and PDS-TD3 are close and both of them outperform PDS-Q-Prop, we use PDS-DDPG in the following due to its simplicity.

\begin{figure}[!htb]
	\vspace{-3mm}
	\centering	
	\subfigure[Average transmission rate in each TF.]{
		\label{fig:constant_policy} 
		\includegraphics[width=0.487\textwidth]{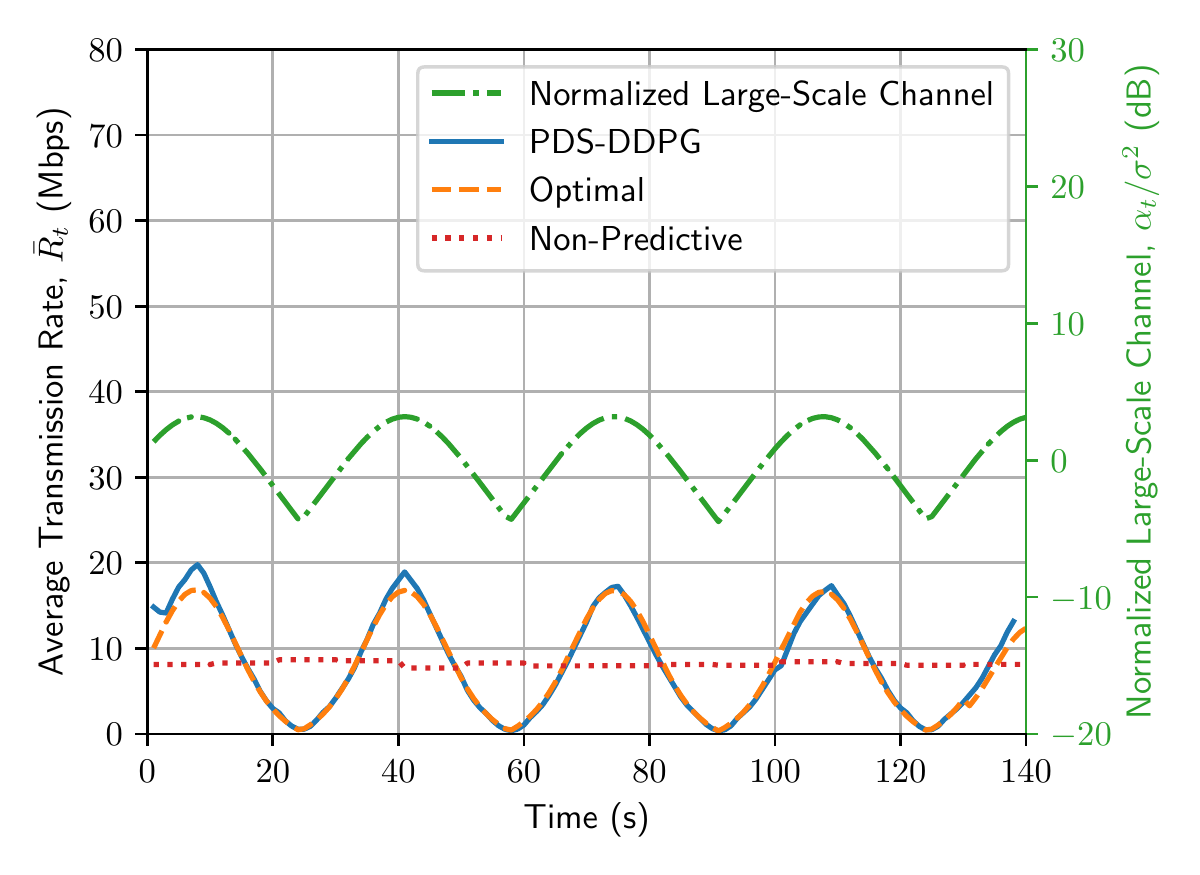}}
	\subfigure[Power allocation]{
		\label{fig:constant_power} 
		\includegraphics[width=0.487\textwidth]{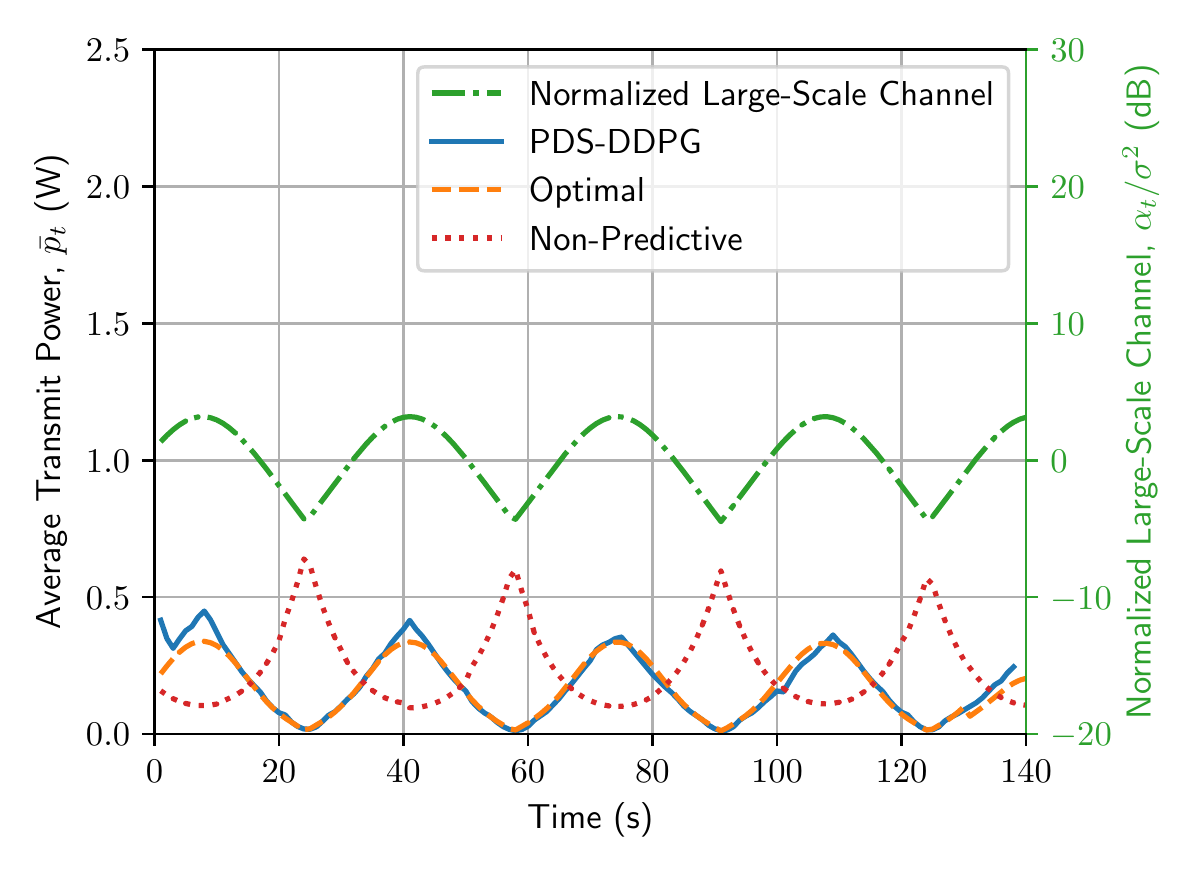}}
	\caption{Policy behavior comparison when users move along the same road with constant velocity.}
	\label{fig:constant}
	\vspace{-2mm}
\end{figure}
In Fig. \ref{fig:constant}, we compare how the proposed policy and the baseline policies behave over time. The result is obtained from an episode after Algorithm~2 has converged. Observe from Fig.~\ref{fig:constant_policy} that the large-scale channel gains vary periodically due to the change of user-to-BS distance as the user moves along the road. Without using any future information, the non-predictive policy has to maintain a constant average transmission rate during the playback of each segment for satisfying the QoS constraint. By contrast, the DRL-based policy behaves similarly to the optimal policy, which transmits more data when the large-scale channel gain is higher. In Fig.~\ref{fig:constant_power}, we compare the energy consumptions of different policies. To maintain a constant average transmission rate, the non-predictive policy has to increase the transmit power in order to compensate the decrease of large-scale channel gain, which results in higher energy consumption. By contrast, the DRL-based policy and the optimal policy allocate less power when the large-scale channel gain decreases, because more data have been transmitted to the user's buffer in advance when the large-scale channel gains are higher.

\subsubsection{Multiple Road \& Random Acceleration}
To show the applicability of the DRL-based policy in more complex scenarios, we consider the scenario of Fig.~\ref{fig:layout2}, where the initial location of each user is randomly chosen from two roads at different distances from the BSs, and the users travel with random acceleration. The initial velocity of each user is randomly chosen within the legitimate velocity range $[10, 20]$ m/s and each user's acceleration in each TF is drawn from the Gaussian distribution with zero mean and standard deviation of $0.3$ m/s$^2$. In this case, the future average channel gains cannot be perfectly predicted due to the random acceleration. Therefore, we compare the proposed DRL-based policy with ``\emph{LSTM \& Optimize}".

\begin{figure}[!htb]
	\vspace{-1mm}
	\centering	
	\subfigure[CDF of energy consumption, the 1st road.]{
		\label{fig:cdf_100m} 
		\includegraphics[width=0.487\textwidth]{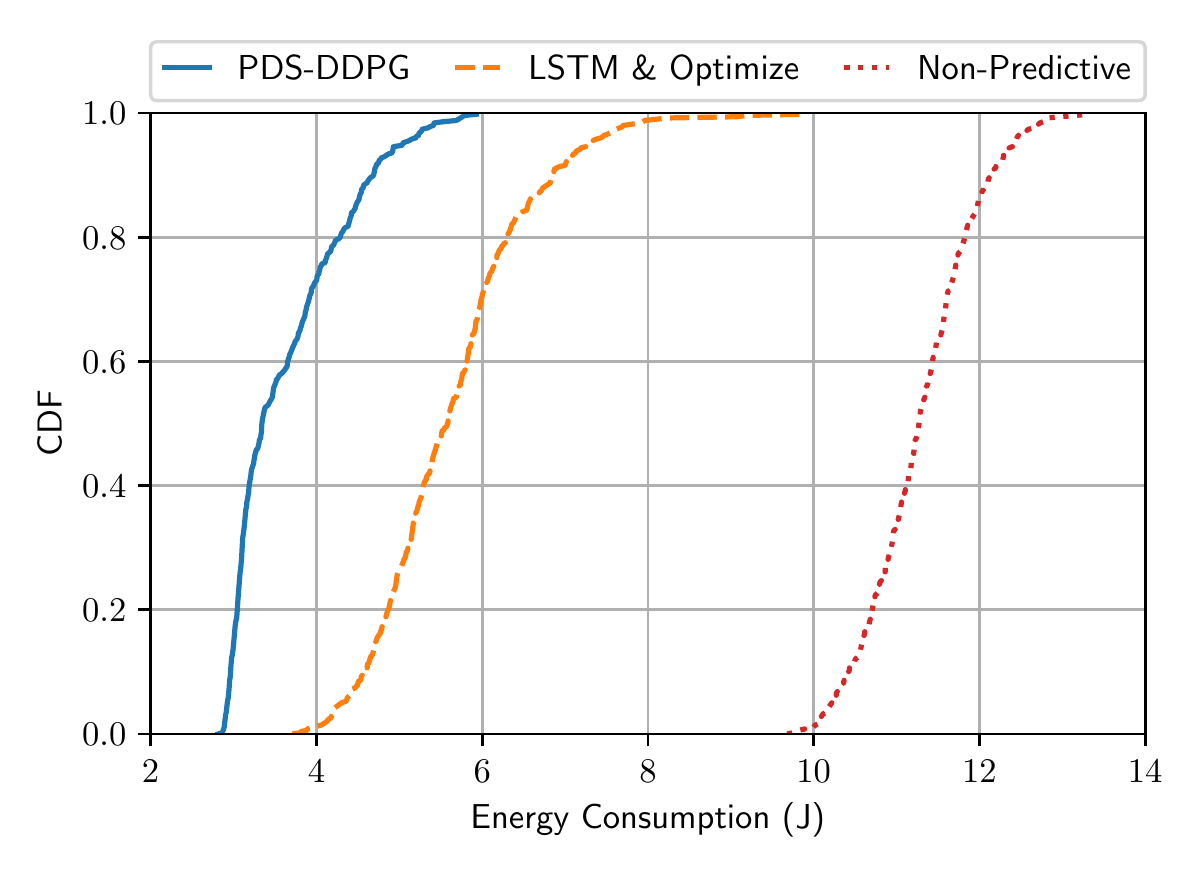}}
	\subfigure[CDF of energy consumption, the 2nd road.]{
		\label{fig:cdf_200m} 
		\includegraphics[width=0.487\textwidth]{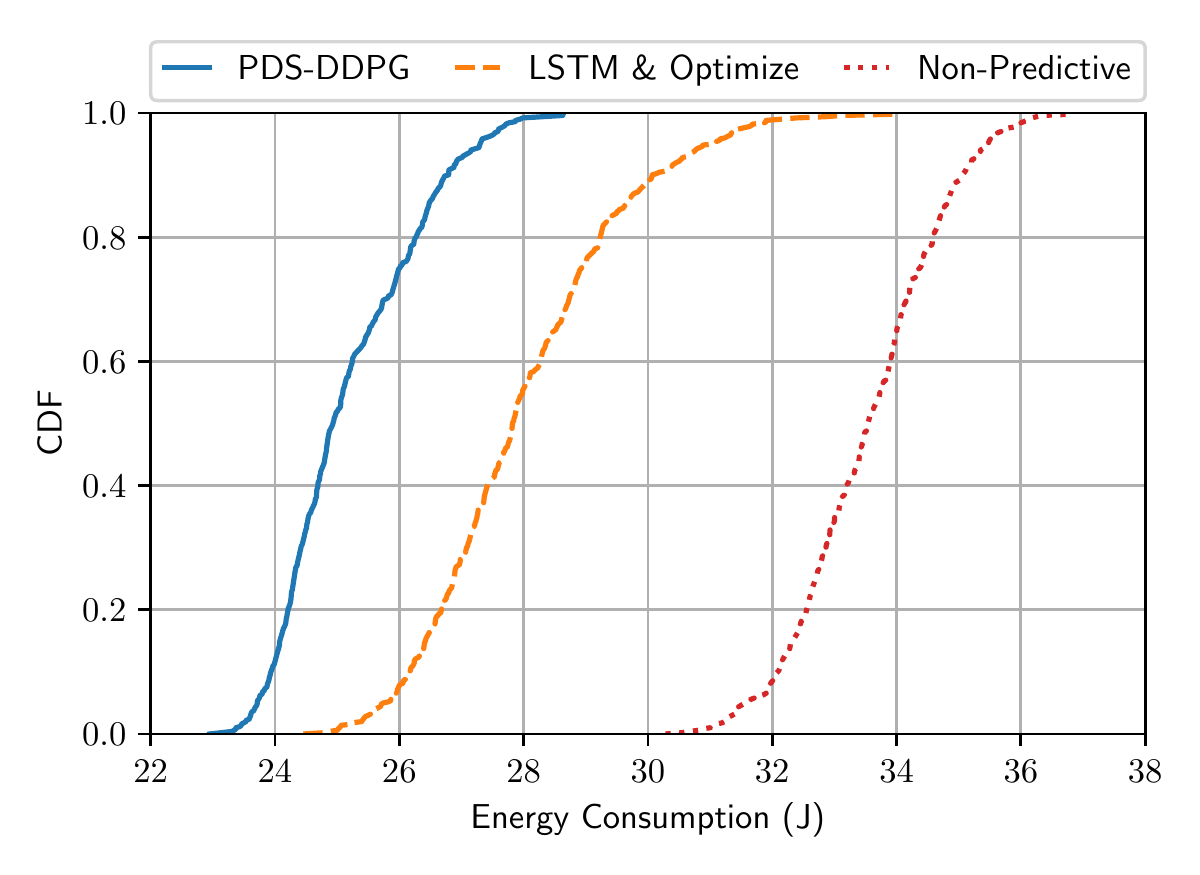}}	
	\subfigure[Policy behavior, the 1st road, initial speed $11$ m/s.]{
		\label{fig:100m} 
		\includegraphics[width=0.487\textwidth]{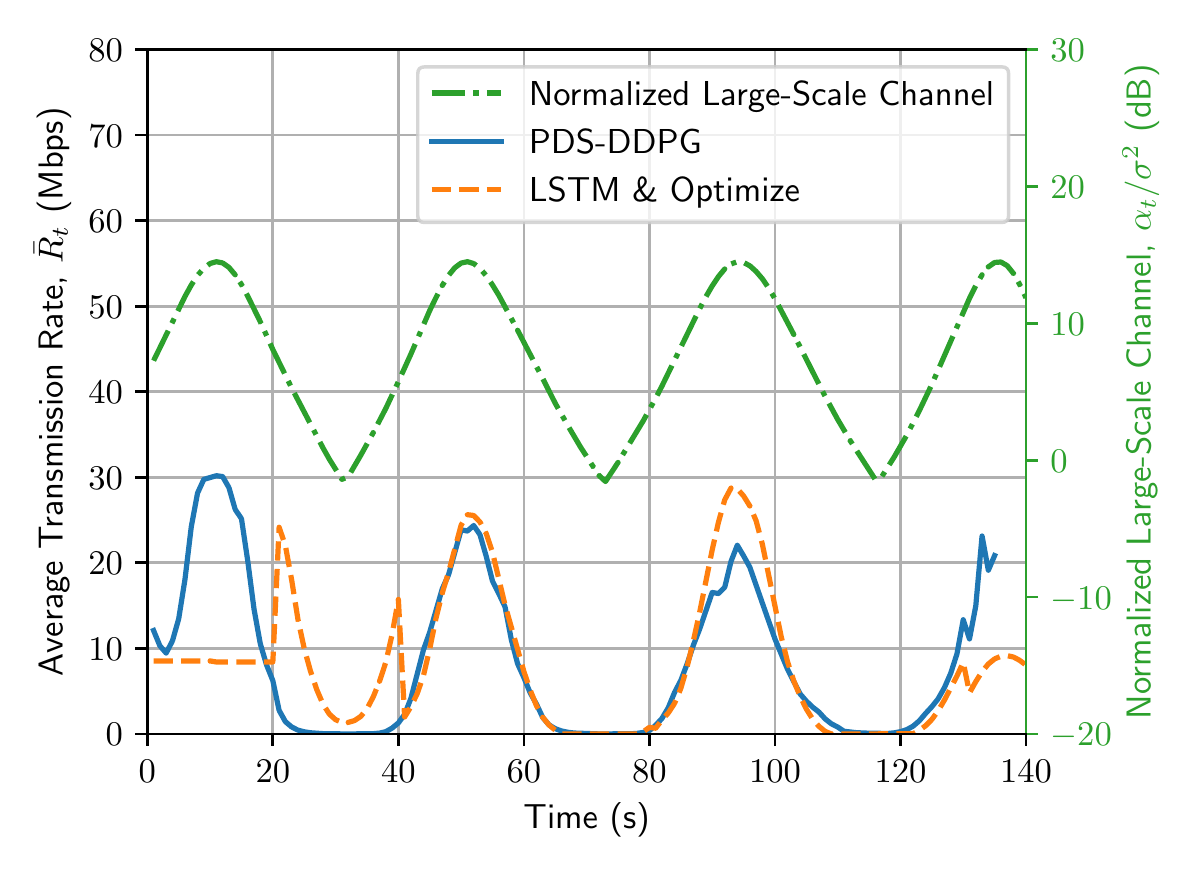}}
	\subfigure[Policy behavior, the 2nd road, initial speed $18$ m/s.]{
		\label{fig:200m} 
		\includegraphics[width=0.487\textwidth]{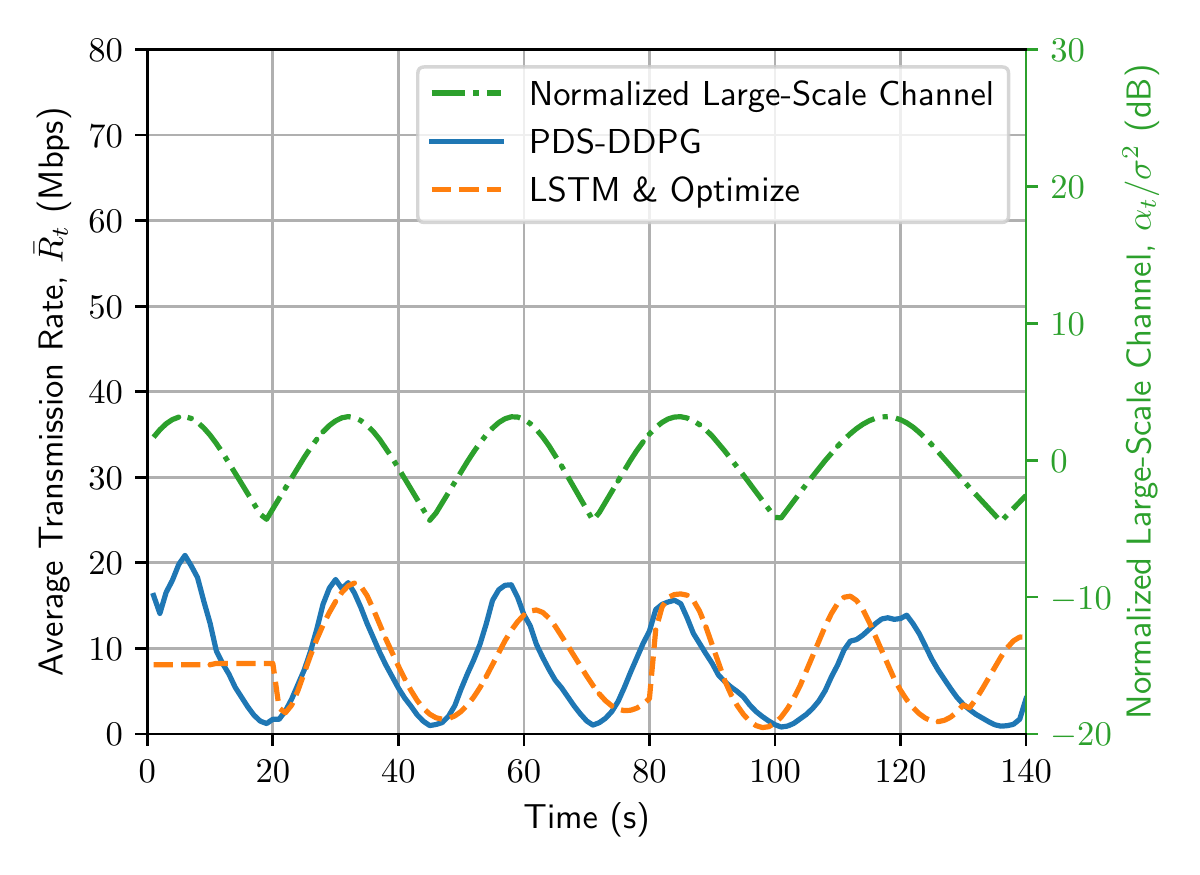}}	
	\caption{Performance and policy behavior comparison when users move along multiple roads with random acceleration.}
	\label{fig:100200}
\end{figure}

In Fig.~\ref{fig:100200}, we compare the performance and the policy behavior. Observe from Fig.~\ref{fig:cdf_100m} and \ref{fig:cdf_200m} that the DRL-based policy achieves lower energy consumption than ``\emph{\textcolor{blue}{LSTM} \& Optimize}" on both roads. This is because the pre-determined average transmission rate of ``\emph{\textcolor{blue}{LSTM} \& Optimize}" after obtaining the predicted information cannot promptly adapt to the real evolution of large-scale channel gains due to the user's random acceleration, as shown in Fig.~\ref{fig:100m} and \ref{fig:200m}. \textcolor{blue}{Moreover, ``\emph{LSTM \& Optimize}" requires an observation window (i.e., the first $20$ s) to gather the information required for predicting the future channel gains. During the observation window, the BS has to serve the user using non-predictive transmission.}  By contrast, PDS-DDPG learns a policy that can adjust the average data rate on-line in order to adapt to the channel variations for both roads \textcolor{blue}{with different initial velocities}. 

\subsubsection{Random Stop}
Let us now consider the scenario of Fig.~\ref{fig:layout3}, where the users may encounter a traffic light on the road. In this scenario, the initial locations of users are uniformly distributed along the road and they may stop for $0\sim 60$ s upon encountering a red traffic light. Since the instant of when and the duration of how long the user stops for are random, it is much harder to predict the future channels for a minute-long prediction window.

\begin{figure}[!htb]
	\vspace{-3mm}
	\centering	
	\subfigure[CDF of energy consumption.]{
		\label{fig:cdf_stop} 
		\includegraphics[width=0.487\textwidth]{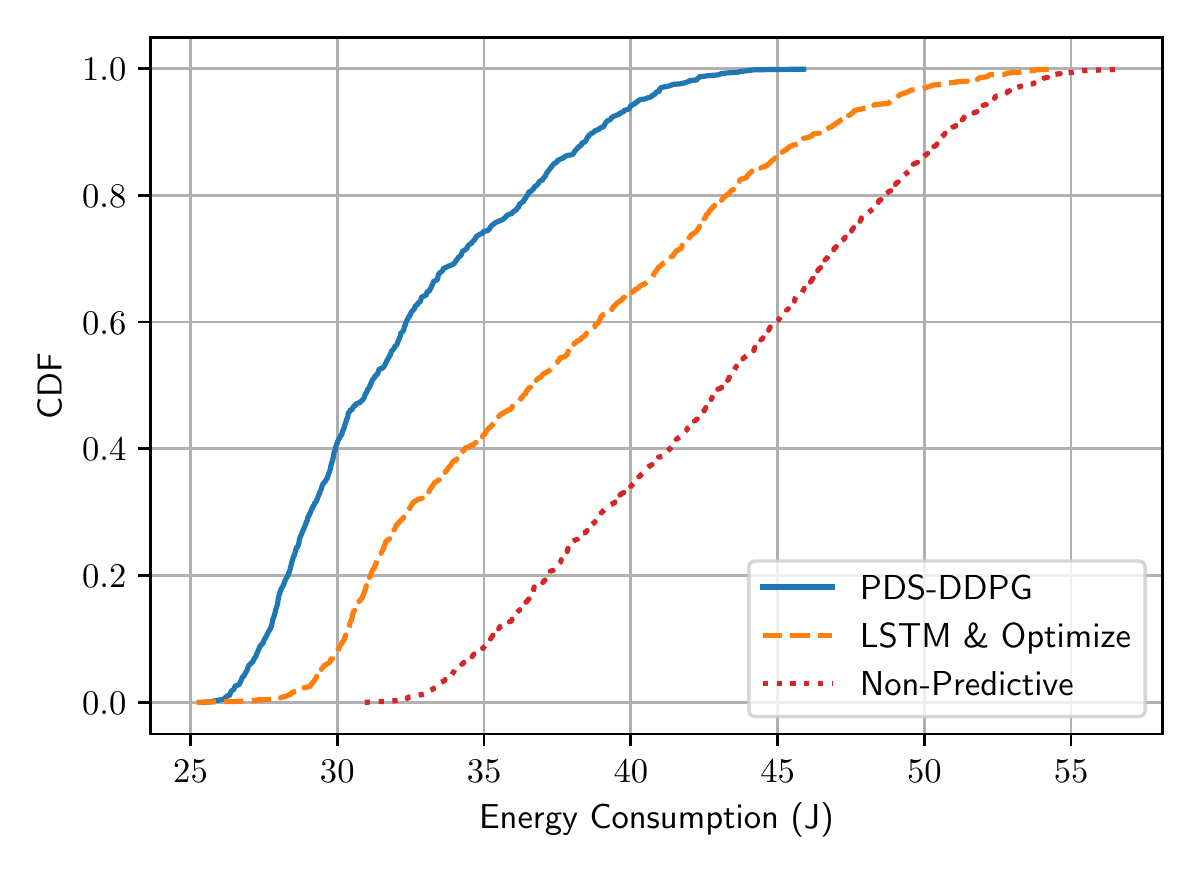}}
	\subfigure[Policy behavior.]{
		\label{fig:stop_policy} 
		\includegraphics[width=0.487\textwidth]{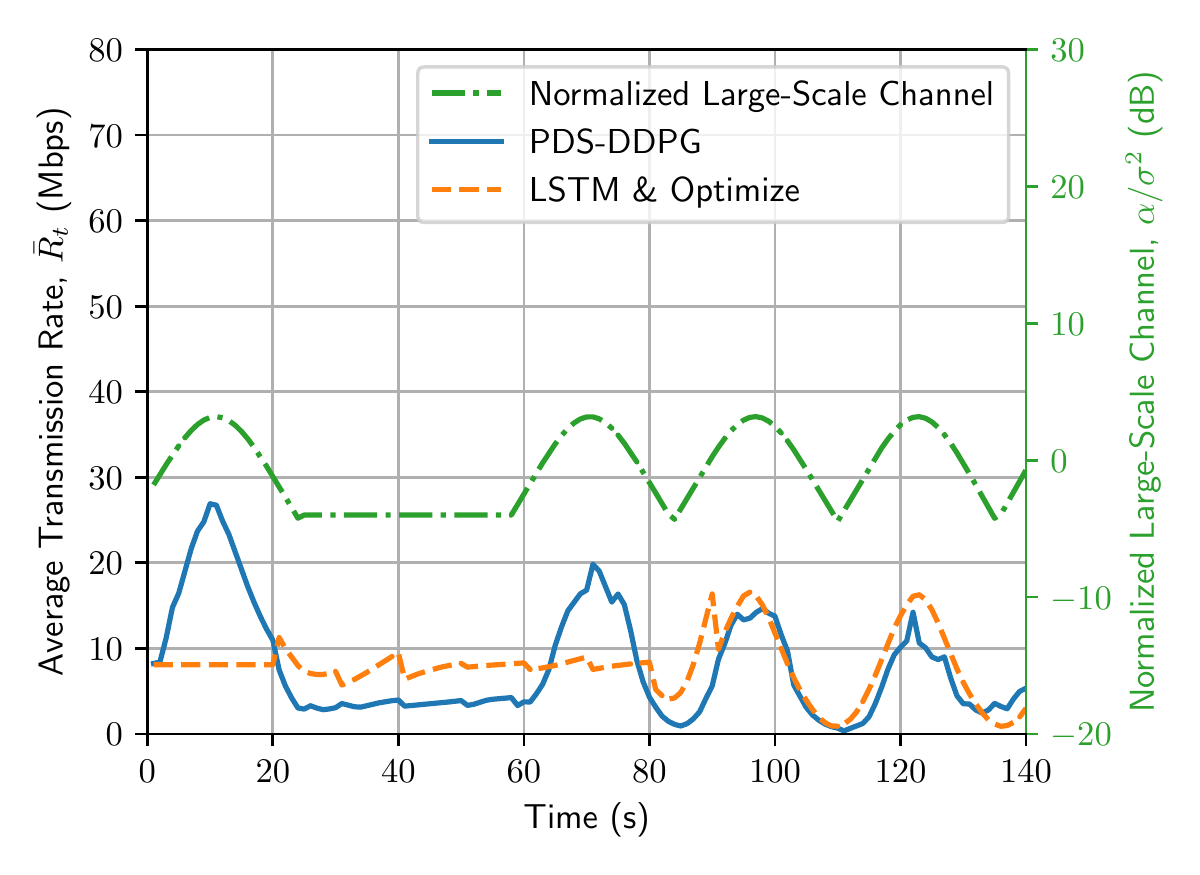}}\vspace{-2mm}
	\caption{Performance and policy behavior comparison when users may stop with a random duration.}
	\label{fig:stop}
\end{figure}

 Again,  we can see from Fig.~\ref{fig:cdf_stop} that PDS-DDPG outperforms ``\emph{LSTM\& Optimize}". In Fig. \ref{fig:stop_policy}, we show a representative episode to compare the policies' behavior, where the user stops at $24$ s and starts to move again at $58$ s. ``\emph{LSTM \& Optimize}" cannot promptly adapt to the channel evolution during $60\sim 80$ s because its average transmission rate for $20\sim 80$ is pre-determined at $20$ s (the end of the observation window) based on the inaccurate channel prediction.  By contrast, as an on-line approach, PDS-DDPG can learn a good policy to adapt to the channel variations.

\section{Conclusions}
In this paper, we proposed a DRL-based policy for optimizing predictive power allocation for video streaming over mobile networks aimed at minimizing the average energy consumption under the QoS constraint. We resorted to DDPG to learn the policy. To reduce the signaling overhead between the MEC server and each BS, we judiciously designed the action and the state by exploiting the knowledge of small-scale fading distribution. To guarantee the QoS constraint during learning and improve the sample efficiency, we integrated the concepts of
safety layer, post-decision state, and virtual experiences into the basic DDPG algorithm by exploiting the partially known model of the system. We have also shown when those accelerating techniques can be extended to other wireless tasks and can be implemented upon other DRL algorithms. Our simulation results have shown that the proposed policy can converge to the optimal policy that is derived based on perfect future large-scale channel gains. When prediction errors exist, the proposed policy outperforms the \emph{first-predict-then-optimize} policy. By exploiting the partially known model, the convergence speed can be dramatically improved.

\bibliographystyle{IEEEtran}

\vspace{-1mm}
\bibliography{dongbib}

\begin{thebibliography}{10}
\providecommand{\url}[1]{#1}
\csname url@samestyle\endcsname
\providecommand{\newblock}{\relax}
\providecommand{\bibinfo}[2]{#2}
\providecommand{\BIBentrySTDinterwordspacing}{\spaceskip=0pt\relax}
\providecommand{\BIBentryALTinterwordstretchfactor}{4}
\providecommand{\BIBentryALTinterwordspacing}{\spaceskip=\fontdimen2\font plus
\BIBentryALTinterwordstretchfactor\fontdimen3\font minus
  \fontdimen4\font\relax}
\providecommand{\BIBforeignlanguage}[2]{{%
\expandafter\ifx\csname l@#1\endcsname\relax
\typeout{** WARNING: IEEEtran.bst: No hyphenation pattern has been}%
\typeout{** loaded for the language `#1'. Using the pattern for}%
\typeout{** the default language instead.}%
\else
\language=\csname l@#1\endcsname
\fi
#2}}
\providecommand{\BIBdecl}{\relax}
\BIBdecl

\bibitem{dongGC19}
D.~Liu, J.~Zhao, and C.~Yang, ``Energy-saving predictive video streaming with
  deep reinforcement learning,'' in \emph{Proc. Globecom}, Waikoloa, USA, 2019,
  pp. 1--6.

\bibitem{index2017global}
Cisco, ``Global mobile data traffic forecast update, 2017--2022 white paper,''
  Feb. 2018.

\bibitem{zhang2018trajectory}
W.~Zhang, Y.~Liu, T.~Liu, and C.~Yang, ``Trajectory prediction with recurrent
  neural networks for predictive resource allocation,'' in \emph{Proc. IEEE
  ICSP}, Beijing, China, 2018, pp. 634--639.

\bibitem{kasparick2015kernel}
M.~Kasparick, R.~L. Cavalcante, S.~Valentin, S.~Sta{\'n}czak, and M.~Yukawa,
  ``Kernel-based adaptive online reconstruction of coverage maps with side
  information,'' \emph{IEEE Trans. Veh. Technol.}, vol.~65, no.~7, pp.
  5461--5473, Jul. 2015.

\bibitem{tsilimantos2016anticipatory}
D.~Tsilimantos, A.~Nogales-G{\'o}mez, and S.~Valentin, ``Anticipatory radio
  resource management for mobile video streaming with linear programming,'' in
  \emph{Proc. IEEE ICC}, Kuala Lumpur, Malaysia, 2016, pp. 1--6.

\bibitem{abou2014energy}
H.~Abou-Zeid, H.~S. Hassanein, and S.~Valentin, ``Energy-efficient adaptive
  video transmission: Exploiting rate predictions in wireless networks,''
  \emph{IEEE Trans. Veh. Technol.}, vol.~63, no.~5, pp. 2013--2026, Jun. 2014.

\bibitem{atawia2017robust}
R.~Atawia, H.~S. Hassanein, H.~Abou-Zeid, and A.~Noureldin, ``Robust content
  delivery and uncertainty tracking in predictive wireless networks,''
  \emph{IEEE Trans. Wireless Commun.}, vol.~16, no.~4, pp. 2327--2339, Apr.
  2017.

\bibitem{she2015context}
C.~She and C.~Yang, ``Context aware energy efficient optimization for video
  on-demand service over wireless networks,'' in \emph{Proc. IEEE/CIC ICCC},
  Shenzhen, China, 2015, pp. 1--6.

\bibitem{mobility}
N.~Bui and J.~Widmer, ``Data-driven evaluation of anticipatory networking in
  {LTE} networks,'' \emph{IEEE Trans. on Mobile Comput.}, vol.~17, no.~10, pp.
  2252--2265, Oct. 2018.

\bibitem{GY18}
J.~Guo, C.~Yang, and C.-L. I, ``Exploiting future radio resources with
  end-to-end prediction by deep learning,'' \emph{IEEE Access}, vol.~6, Dec.
  2018.

\bibitem{scy}
C.~{She} and C.~{Yang}, ``Energy efficient resource allocation for hybrid
  services with future channel gains,'' \emph{IEEE Trans. Green Commun. and
  Netw.}, vol.~4, no.~1, pp. 165--179, Mar. 2020.

\bibitem{sutton1998reinforcement}
R.~S. Sutton and A.~G. Barto, \emph{Reinforcement learning: An
  introduction}.\hskip 1em plus 0.5em minus 0.4em\relax MIT press Cambridge,
  1998.

\bibitem{lecun2015deep}
Y.~LeCun, Y.~Bengio, and G.~Hinton, ``Deep learning,'' \emph{Nature}, vol. 521,
  no. 7553, pp. 436--444, 2015.

\bibitem{hu2015mobile}
Y.~C. Hu, M.~Patel, D.~Sabella, N.~Sprecher, and V.~Young, ``Mobile edge
  computing: A key technology towards {5G},'' \emph{ETSI white paper}, vol.~11,
  no.~11, pp. 1--16, 2015.

\bibitem{DRL}
N.~C. {Luong}, D.~T. {Hoang}, S.~{Gong}, D.~{Niyato}, P.~{Wang}, Y.~{Liang},
  and D.~I. {Kim}, ``Applications of deep reinforcement learning in
  communications and networking: A survey,'' \emph{IEEE Commun. Surveys Tuts.},
  vol.~21, no.~4, pp. 3133--3174, Fourthquarter 2019.

\bibitem{zhao2019deep}
N.~{Zhao}, Y.~{Liang}, D.~{Niyato}, Y.~{Pei}, M.~{Wu}, and Y.~{Jiang}, ``Deep
  reinforcement learning for user association and resource allocation in
  heterogeneous cellular networks,'' \emph{IEEE Trans. Wireless Commun.},
  vol.~18, no.~11, pp. 5141--5152, Nov. 2019.

\bibitem{zhang2019proactive}
Z.~{Zhang}, Y.~{Yang}, M.~{Hua}, C.~{Li}, Y.~{Huang}, and L.~{Yang},
  ``Proactive caching for vehicular multi-view {3D} video streaming via deep
  reinforcement learning,'' \emph{IEEE Trans. Wireless Commun.}, vol.~18,
  no.~5, pp. 2693--2706, May 2019.

\bibitem{liu2019DRL}
D.~{Liu} and C.~{Yang}, ``A deep reinforcement learning approach to proactive
  content pushing and recommendation for mobile users,'' \emph{IEEE Access},
  vol.~7, pp. 83\,120--83\,136, 2019.

\bibitem{DDPG}
T.~P. Lillicrap, J.~J. Hunt, A.~Pritzel, N.~Heess, T.~Erez, Y.~Tassa,
  D.~Silver, and D.~Wierstra, ``Continuous control with deep reinforcement
  learning,'' in \emph{Proc. ICLR}, San Juan, Puerto Rico, 2016, pp. 1--14.

\bibitem{dalal2018safe}
\BIBentryALTinterwordspacing
G.~Dalal, K.~Dvijotham, M.~Vecerik, T.~Hester, C.~Paduraru, and Y.~Tassa,
  ``Safe exploration in continuous action spaces,'' \emph{arXiv 1801.08757
  [cs.AI]}, 2018. [Online]. Available: \url{http://arxiv.org/abs/1801.08757}
\BIBentrySTDinterwordspacing

\bibitem{mastronarde2011fast}
N.~Mastronarde and M.~van~der Schaar, ``Fast reinforcement learning for
  energy-efficient wireless communication,'' \emph{IEEE Trans. Signal
  Process.}, vol.~59, no.~12, pp. 6262--6266, Dec. 2011.

\bibitem{fujimoto2018addressing}
S.~Fujimoto, H.~Hoof, and D.~Meger, ``Addressing function approximation error
  in actor-critic methods,'' in \emph{Proc. ICML}, Stockholm, Sweden, 2018, pp.
  1582--1591.

\bibitem{gu2016q}
S.~Gu, T.~Lillicrap, Z.~Ghahramani, R.~E. Turner, and S.~Levine, ``{Q-{P}rop}:
  Sample-efficient policy gradient with an off-policy critic,'' in \emph{Proc.
  ICLR}, Toulon, France, 2017, pp. 1--13.

\bibitem{giust2018mec}
F.~Giust, G.~Verin, K.~Antevski, J.~Chou, Y.~Fang, W.~Featherstone, F.~Fontes,
  D.~Frydman, A.~Li, A.~Manzalini \emph{et~al.}, ``{MEC} deployments in {4G}
  and evolution towards {5G},'' \emph{ETSI White Paper}, vol.~24, pp. 1--24,
  2018.

\bibitem{energy}
G.~{Auer}, V.~{Giannini}, C.~{Desset}, I.~{Godor}, P.~{Skillermark},
  M.~{Olsson}, M.~A. {Imran}, D.~{Sabella}, M.~J. {Gonzalez}, O.~{Blume}, and
  A.~{Fehske}, ``How much energy is needed to run a wireless network?''
  \emph{IEEE Wireless Commun.}, vol.~18, no.~5, pp. 40--49, Oct. 2011.

\bibitem{mnih2015human}
V.~Mnih, K.~Kavukcuoglu, D.~Silver \emph{et~al.}, ``Human-level control through
  deep reinforcement learning,'' \emph{Nature}, vol. 518, no. 7540, p. 529,
  Feb. 2015.

\bibitem{youtube}
Youtube, ``Recommended upload encoding setting,''
  \url{https://support.google.com/youtube/answer/1722171}, {Accessed}: Jul.
  2020.

\bibitem{tensorflow2015-whitepaper}
\BIBentryALTinterwordspacing
M.~Abadi, A.~Agarwal, P.~Barham, E.~Brevdo \emph{et~al.}, ``{TensorFlow}:
  Large-scale machine learning on heterogeneous systems,'' 2015, software
  available from tensorflow.org. [Online]. Available:
  \url{http://tensorflow.org/}
\BIBentrySTDinterwordspacing

\bibitem{adam}
D.~P. Kingma and J.~Ba, ``Adam: A method for stochastic optimization,'' in
  \emph{Proc. ICLR}, Banff, Canada, 2014, pp. 1--15.

\bibitem{baselines}
P.~Dhariwal, C.~Hesse, O.~Klimov, A.~Nichol, M.~Plappert, A.~Radford,
  J.~Schulman, S.~Sidor, Y.~Wu, and P.~Zhokhov, ``Open{AI} baselines,''
  \url{https://github.com/openai/baselines}, 2017.

\bibitem{SpinningUp2018}
J.~Achiam, ``Spinning up in deep reinforcement learning,''
  \url{https://github.com/openai/spinningup}, 2018.

\bibitem{GAE}
J.~Schulman, P.~Moritz, S.~Levine, M.~Jordan, and P.~Abbeel, ``High-dimensional
  continuous control using generalized advantage estimation,'' in \emph{Proc.
  ICLR}, San Juan, Puerto Rico, 2016, pp. 1--14.

\bibitem{chollet2015keras}
F.~Chollet \emph{et~al.}, ``Keras,'' \url{https://github.com/fchollet/keras},
  2015.

\end{thebibliography}

\appendices

\section{Proof of Proposition 1}
\renewcommand{\theequation}{A.\arabic{equation}}
\setcounter{equation}{0}
In the following, we first rewrite problem $\sf P2$ as a functional optimization problem.
Since the large-scale channel gain remains constant within a TF and the power allocation policy within a TF should adapt to the small-scale fading, $p_{ti}$ can be expressed as a function of $g_{ti}$ as $p_{ti} = p(g_{ti})$. Considering that $g_{ti}$ is i.i.d. among TSs, \eqref{aver-enegry} can be rewritten as
\begin{equation}
\bar E_t =\mathbb{E}_{g_{ti}}\left[ \frac{1}{\rho_{\rm E}}\sum_{i=1}^{N_{\rm s}} \tau p (g_{ti}) \right] + \Delta T P_{\rm c} = \Delta T \left( \frac{1}{\rho_{\rm E}} \mathbb E_{g_{ti}} \left[p(g_{ti})\right] + P_{\rm c} \right). \label{eqn:Et}
\end{equation}
Since the second term $P_{\rm c}$ in \eqref{eqn:Et} does not depend on the power allocation ${p}(g_{ti})$, problem ${\sf P2}$ is equivalent to the following functional optimization problem.
\begin{subequations}
	\begin{align}
	{\sf P3}:\forall t, \quad \min_{p(g_{ti})} ~ & \mathbb{E} \left[p(g_{ti})\right] \\
	s.t. ~& \mathbb{E}_{g_{ti}}\left[W\log_2\left(1 + \frac{\alpha_t}{\sigma^2}p(g_{ti}) g_{ti}\right)\right] = \bar R_t \label{eqn:con1}\\
	& 0 \leq p(g_{ti}) \leq P_{\max}, ~\forall g_{i}. \label{eqn:con2}
	\end{align}
\end{subequations}

The Lagrangian function of problem $\sf P3$ can be expressed as
\begin{align}
\!\!&\mathcal{L}(p(g_{ti}), \lambda_1(g_{ti}), \lambda_2(g_{ti}), \mu_t)=\nonumber \\
\!\!& \mathbb{E}_{g_{ti}}\left[ p(g_{ti}) - \lambda_1(g_{ti})p(g_{ti})  + \lambda_2(g_{ti})(p(g_{ti})\! - \!P_{\max}) + \mu_t \! \left(\bar R_t - W \log_2\left(1 + \tfrac{\alpha_t}{\sigma^2}p(g_{ti}) g_{ti}\right)\right)  \right], \!\! \nonumber
\end{align}
where $\lambda_1(g_{ti}), \lambda_2(g_{ti})$ and $\mu_t$ are the multipliers associated with the inequality and equality constraints, respectively.
The Karush-Kuhn-Tucker (KKT) conditions of problem ${\sf P3}$ are:
\begin{subequations} \label{eqn:KKT}
	\begin{align}
	\frac{\partial\mathcal{L}}{\partial p(g_{ti})} = \mathbb{E}_{g_{ti}}\left[ 1 - \lambda_1(g_{ti}) + \lambda_2(g_{ti}) - \frac{\xi_t}{\sigma^2(\alpha_tg_{ti})^{-1} + p(g_{ti})}\right]  = 0 & \label{eqn:sta}\\
	\lambda_1(g_{ti})p(g_{ti}) = 0&, ~\forall g_{ti} \label{eqn:com1}\\
	\lambda_2(g_{ti})(p(g_{ti})-P_{\max})  = 0&, ~\forall g_{ti} \label{eqn:com2}\\
	\eqref{eqn:con1}, \eqref{eqn:con2}, \lambda_1(g_{ti}) \geq 0, \lambda_2(g_{ti})\geq 0&,  ~ \forall g_{ti}
	\end{align}
\end{subequations}
where $\xi_t = \frac{\mu_t W}{\ln 2}$, and the stationary condition \eqref{eqn:sta} can be simplified to
\begin{equation}
1 - \lambda_1(g_{ti}) + \lambda_2(g_{ti}) -  \frac{\xi_t}{\sigma^2(\alpha_tg_{ti})^{-1} + p(g_{ti})} = 0. \label{eqn:sta2}
\end{equation}

In what follows, we find the solution of \eqref{eqn:KKT}.
We first prove $\xi_t > 0$. Assuming that $\xi_t \leq 0$, we can obtain $\lambda_1(g_{ti}) \geq 1 + \lambda_2(g_{ti})$ according to \eqref{eqn:sta2}. Then, since $\lambda_2(g_{ti})\geq 0$, we have $\lambda_1(g_{ti}) \geq 0$. Based on \eqref{eqn:com1}, we have $p(g_{ti}) = 0, \forall g_{ti}$, which contradicts  to \eqref{eqn:con1}. Therefore, we have $\xi_t > 0$.

Next, we derive $p(g_{ti})$ under different conditions.
When $g_{ti}<\frac{\sigma^2}{\alpha_t \xi_t}$, we have $1 - \frac{\xi_t}{\sigma^2(\alpha_tg_{ti})^{-1} + p(g_{ti})}> 0$. In this case, according to \eqref{eqn:sta2}, we can obtain $\lambda_1(g_{ti}) > \lambda_2(g_{ti})$. Further considering that $\lambda_2(g_{ti})\geq 0$, we have $\lambda_1 (g_{ti}) > 0$. Then, according to \eqref{eqn:com1}, we obtain $p(g_{ti}) = 0$. When $g_{ti} = \frac{\sigma^2}{\alpha_t \xi_t}$, we have $1 - \frac{\xi_t}{\sigma^2(\alpha_tg_{ti})^{-1} + p(g_{ti})}> 0$ if $p(g_{ti}) > 0$. However, from $1 - \frac{\xi_t}{\sigma^2(\alpha_tg_{ti})^{-1} + p(g_{ti})}> 0$, we can obtain $p(g_{ti}) = 0$ again based on \eqref{eqn:sta2} and $\lambda_2(g_{ti})\geq 0$, which contradicts to $p(g_{ti}) > 0$. Therefore, we have $p(g_{ti}) = 0$.

When $g_{ti} > \frac{\sigma^2}{\alpha_t(\xi_t - P_{\max})}$, we have $1 - \frac{\xi_t}{\sigma^2(\alpha_tg_{ti})^{-1} + p(g_{ti})} < 0$. In this case, according to \eqref{eqn:sta2}, we can obtain $\lambda_2(g_{ti}) > \lambda_1(g_{ti})$. Further considering $\lambda_1(g_{ti})\geq 0$, we have $\lambda_2 (g_{ti}) > 0$. Then, according to \eqref{eqn:com2}, we can obtain $p(g_{ti}) = P_{\max}$. When $g_{ti} = \frac{\sigma^2}{\alpha_t(\xi_t - P_{\max})}$, we have $1 - \frac{\xi_t}{\sigma^2(\alpha_tg_{ti})^{-1} + p(g_{ti})} < 0$ if $p(g_{ti}) < P_{\max}$. However, from $1 - \frac{\xi_t}{\sigma^2(\alpha_tg_{ti})^{-1} + p(g_{ti})} < 0$ we can obtain $p(g_{ti}) = P_{\max}$ again based on \eqref{eqn:sta2} and $\lambda_1(g_{ti})\geq 0$, which contradicts to $p(g_{ti}) < P_{\max}$. Therefore, $p(g_{ti}) = P_{\max}$.

When $\frac{\sigma^2}{\alpha_t \xi_t} < g_{ti} < \frac{\sigma^2}{\alpha_t(\xi_t - P_{\max})}$, we have $\frac{\sigma^2}{\alpha_t g_{ti}}  <\xi_t < \frac{\sigma^2}{\alpha_t g_{ti}} + P_{\max} $. In this case, if $p(g_{ti}) = 0$, according to \eqref{eqn:sta2}, we can obtain $\lambda_2(g_{ti})>\lambda_1(g_{ti})$. Further considering that $\lambda_1(g_{ti})\geq 0$, we have $\lambda_2 (g_{ti}) > 0$. Then, according to \eqref{eqn:com2}, we have $p(g_{ti}) = P_{\max}$, which contradicts to $p(g_{ti}) = 0$ and hence $p(g_{ti}) > 0$. Similarity, if $p(g_{ti}) = P_{\max}$, according to \eqref{eqn:sta2}, we can obtain $\lambda_1(g_{ti})>\lambda_2(g_{ti})$. Further considering that $\lambda_2(g_{ti})\geq 0$, we have $\lambda_1 (g_{ti}) > 0$. Then, according to \eqref{eqn:com1}, we have $p(g_{ti}) = 0$, which contradicts to $p(g_{ti}) = P_{\max}$. Hence, $p(g_{ti}) < P_{\max}$. Therefore, we have $0 < p_{g_{ti}} < P_{\max}$. Consequently, we can obtain $\lambda_1(g_{ti}) = \lambda_2(g_{ti}) =0$. By substituting $\lambda_1(g_{ti}) = \lambda_2(g_{ti}) =0$ into \eqref{eqn:sta2}, we have $p(g_{ti}) = \xi_t - \frac{\sigma^2}{\alpha_t g_{ti}}$.

Finally, by summarizing the above results and further considering the average rate constraint \eqref{eqn:con1}, Proposition 1 is proved. \qed


\section{Proof of Proposition 2}
\renewcommand{\theequation}{B.\arabic{equation}}
\setcounter{equation}{0}
Considering that $\Delta T = N_{\rm s} \tau$, we obtain
$
\sum_{i=1}^{N_{\rm s}} \tau p_{ti} = \sum_{i=1}^{N_{\rm s}} \frac{\Delta T}{N_{\rm s}} p_{ti} =\Delta T \sum_{i=1}^{N_{\rm s}}  \frac{p_{ti}}{N_{\rm s}}
$.
When $\tau \ll \Delta T$, we have $N_{\rm s} = \frac{\Delta T}{\tau} \to \infty$. Since $p_{ti}$ is a function of $g_{ti}$, which is i.i.d. among TSs, we can apply the law of large numbers to obtain $ \sum_{i=1}^{N_{\rm s}}  \frac{p_{ti}}{N_{\rm s}} \overset{a.s.}{\rightarrow} \bar p_t$ and hence $\mathrm{Pr} \big(\sum_{i=1}^{N_{\rm s}} \tau p_{ti} = \Delta T\bar p_t\big) = 1$. Similarity, we can obtain  $\mathrm{Pr} \big(\sum_{i=1}^{N_{\rm s}} \tau R_{ti} = \Delta T \bar R_t\big) = 1$. The average transmit power $\bar p_{t} = \int_{0}^{\infty} p^{\rm opt}(\alpha_t g; \xi^{\rm opt}(\bar R_t)) \rho(g) dg$ can be derived from Proposition 1.
Specifically, for Rayleigh fading and large transmit power, $\bar p_t$ can be expressed as
\begin{align}
\bar p_t & = \int_{\frac{\sigma^2}{\alpha_t\xi^{\rm opt}(\bar R_t)}}^{\infty} \left(\xi^{\rm opt}(\bar R_t) - \frac{\sigma^2}{\alpha_t g}\right) e^{-g} {\rm d}g = \xi_t^{\rm opt}(\bar R_t) e^{-\frac{\sigma^2}{\alpha_t \xi^{\rm opt}(\bar R_t)}} -  \frac{\sigma^2}{\alpha_t} {\rm E}_1 \left( \frac{\sigma^2}{\alpha_t \xi^{\rm opt}(\bar R_t)} \right). \label{eqn:barpt}
\end{align}
By substituting \eqref{eqn:xiopt} into \eqref{eqn:barpt}, Proposition 2 is proved. \qed

\section{Experiment Details}
All experiments are conducted on a work station with AMD Ryzen\texttrademark~9 3950X CPU and a single Nvidia Geforce RTX\texttrademark~2080Ti GPU. The NNs are implemented using TensorFlow 1.15~\cite{tensorflow2015-whitepaper} on Windows 10 and we use Adam \cite{adam} for learning the neural network parameters. 
\subsection{DRL Algorithms}
\subsubsection{DDPG and PDS-DDPG}
\begin{figure}[!htb]
	\vspace{-3mm}
	\centering
	\includegraphics[width=0.487\textwidth]{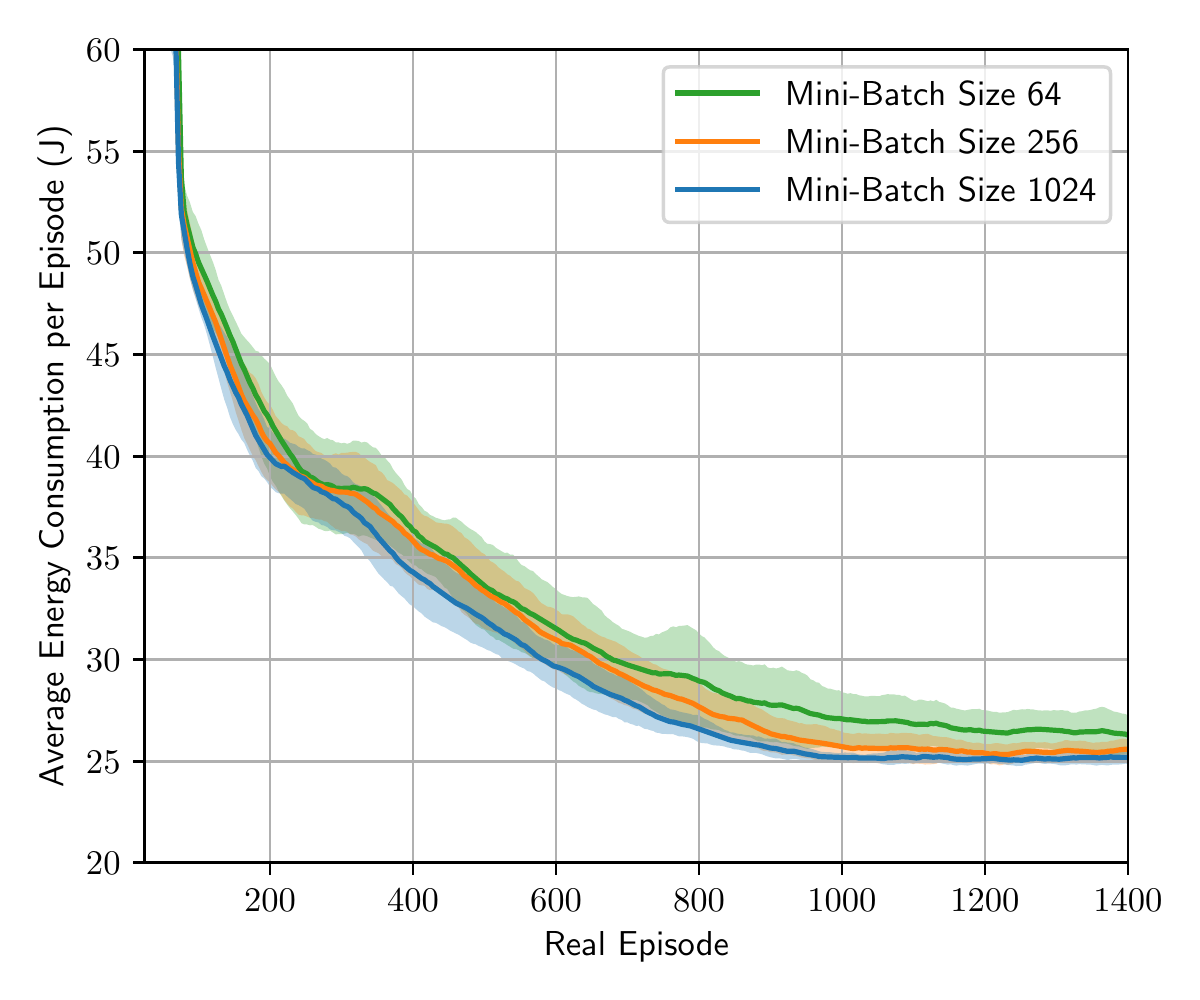}
	\vspace{-4mm}
	\caption{The impact of mini-batch size on PDS-DDPG. To provide a clear figure, we only show the results with mini-batch size $[64, 256, 1024]$. }
	\label{fig:batch}
\end{figure}

Followed by the main settings given in Section V-B, other settings of DDPG and PDS-DDPG are given as follows. For DDPG, the penalty term in \eqref{eqn:reward} is clipped by $[0, 50]$ for stabilizing learning. For both DDPG and PDS-DDPG, the weights of the hidden layers are initialized from uniform distributions $[-\frac{1}{\sqrt{f}}, \frac{1}{\sqrt{f}}]$ according to \cite{DDPG}, where $f$ is the input dimension of the layer, and the biases are initialized as zeros. To ensure the initial outputs of the actor and critic networks close to those of the non-predictive transmission policy, the weights of the actor and critic networks' output layers are initialized from uniform distribution $[-10^{-4}, 10^{-4}]$,  the bias of the actor network's output layer is initialized as $-15$, and the bias of the actor network's output layer is initialized as $-1$. The replay buffer size is $10^6$. The impact of mini-batch size on PDS-DDPG are shown in Fig.~\ref{fig:batch}.

\subsubsection{TD3 and PDS-TD3}
The settings of TD3 and PDS-TD3 are inherited from DDPG and PDS-DDPG, respectively. Moreover, for both TD3 and PDS-TD3, the actor and target critic network's update delay is set as $2$~\cite{fujimoto2018addressing}, the target action addictive noise is sampled from the standard Gaussian distribution and is clipped by $[-10, 10]$.   

\subsubsection{Q-Prop and PDS-Q-Prop}
Our implementations are built based on OpenAI's TRPO~\cite{baselines,SpinningUp2018} with generalized advantage estimation ($\lambda=0.97$)~\cite{GAE}. For Q-Prop, the value network, policy (i.e. actor) network, and Q-network all have two hidden layers with $200$ nodes. For PDS-Q-Prop, the number of hidden nodes can be reduced to $100$ for each hidden layer. The activation functions and initialization method are the same as DDPG and PDS-DDPG. For both Q-Prop and PDS-Q-Prop, we use the conservative version~\cite{gu2016q}. All the networks are updated for every $600$ steps (roughly $4$ episodes). In each update of the value network, those $600$ steps are iterated for $10$ epochs with mini-batch size of $32$ and learning rate of $10^{-3}$. The Q-network are updated in the same way as DDPG and PDS-DDPG for $600$ iterations during each update. We use a Gaussian policy network for both Q-Prop and PDS-Q-Prop, whose output is the mean value of the action and the standard deviation is linearly decreased from $10$ to zero during the training phase (similar to the exploration noise of DDPG and PDS-DDPG). The 
Kullback–Leibler divergence step size for the policy network update is set as $10^{-3}$. For PDS-Q-Prop, the safety layer is applied on the top of the policy network and its Q-network has the same architecture as that of the critic network in PDS-DDPG.  

\subsection{LSTM Network}
The LSTM network in the baseline method ``\emph{LSTM \& Predict}" is configured and trained as follows. We use the many-to-many LSTM architecture built by Keras~\cite{chollet2015keras}. The input features are chosen as the average channel gains (normalized by the noise power) and the user locations (in kilometers) for an observation window of  $20$ s, resulting an input dimension of $2\times 20$. The output of the LSTM is the normalized average channel gains in a prediction window of $60$ s, resulting an output dimension of $1\times 60$. The LSTM has one hidden layer with $200$ nodes. The training set is generated from $200$ user trajectories each with a length of $150$ s, and the LSTM is trained for $10$ epochs with mini-batch size $32$.

\end{document}